\documentclass{article} 
\usepackage{iclr2027_conference,times}

\usepackage{amsmath,amsfonts,bm}

\def\eqref#1{equation~\ref{#1}}

\def\1{\bm{1}}

\DeclareMathAlphabet{\mathsfit}{\encodingdefault}{\sfdefault}{m}{sl}
\SetMathAlphabet{\mathsfit}{bold}{\encodingdefault}{\sfdefault}{bx}{n}

\usepackage{hyperref}
\usepackage{url}

\usepackage{graphicx}
\usepackage{booktabs}       
\usepackage{amsfonts}       
\usepackage{nicefrac}       
\usepackage{xcolor}         

\usepackage{amsmath}
\usepackage{amssymb}
\usepackage{graphicx}
\usepackage{multirow}
\usepackage{bm}
\usepackage{bbm}
\usepackage{array}
\usepackage{makecell}
\usepackage{adjustbox}
\usepackage[table]{xcolor}
\usepackage{pifont}
\usepackage{algorithm}
\usepackage{algpseudocode}
\usepackage{enumitem}

\usepackage{cuted}
\usepackage{wrapfig}
\usepackage{caption}

\usepackage{booktabs}

\usepackage{xcolor}
\definecolor{motionblue}{HTML}{1F6FB2}
\definecolor{motionorange}{HTML}{E2711D}
\definecolor{motiongray}{HTML}{D6DDE8}
\definecolor{motionpink}{HTML}{C71387}

\title{MotionMaestro: Masked Tokenization for Unified Motion Generation}

\author{
Yun Chen \quad Munchurl Kim\thanks{Co-corresponding authors.} \quad Jeonghyeok Do\footnotemark[1]\\
Korea Advanced Institute of Science and Technology (KAIST)\\
\texttt{\{cyruby,mkimee,ehwjdgur0913\}@kaist.ac.kr}\\[4pt]
{\small Project Page: \url{https://kaist-viclab.github.io/MotionMaestro_site/}}
}

\iclrfinalcopy
\begin{document}

\maketitle

\begin{center}
    \vspace{-0.7cm}
    \includegraphics[width=0.9\linewidth]{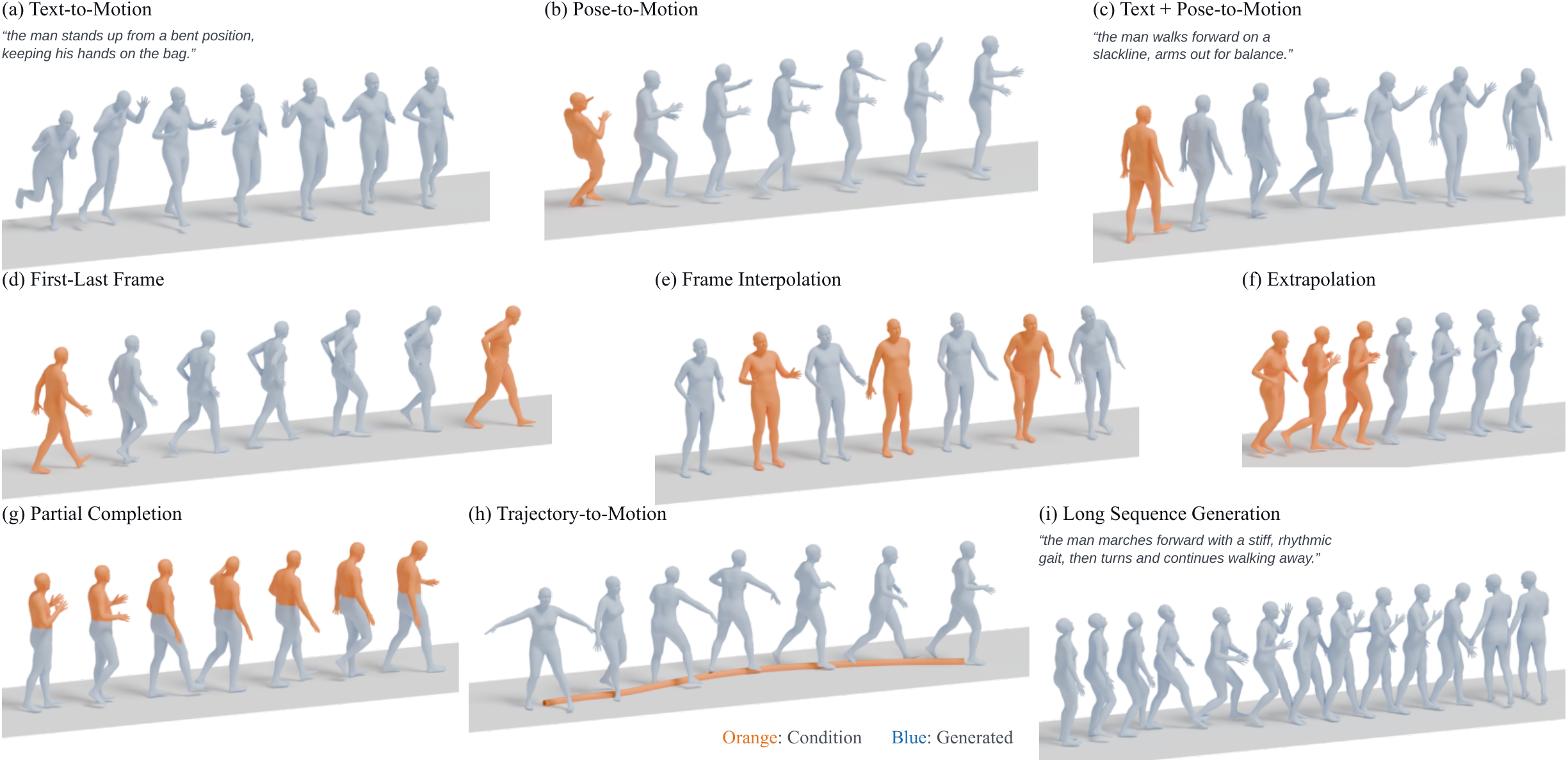}
    \vspace{-0.2cm}
    \captionof{figure}{\textbf{Unified motion generation with MotionMaestro.} MotionMaestro supports various motion generation tasks with a single model: (a) text-to-motion, (b) pose-to-motion, (c) text+pose-to-motion, (d) first-last frame, (e) frame interpolation, (f) extrapolation, (g) partial completion, (h) trajectory-to-motion, and (i) long sequence generation. \textcolor{motionorange}{Orange} denotes provided motion conditions, including observed poses, body parts, and trajectories; \textcolor{motionblue}{blue} denotes generated motion.
}
\label{fig:teaser}
\end{center}

\begin{abstract}
Human motion generation plays an important role in applications such as character animation, virtual environments, and embodied interaction. While existing approaches have achieved remarkable progress, many of them are developed for individual tasks, including text-to-motion, pose-conditioned generation, and trajectory control. Although these tasks involve different types of conditions, a unified framework capable of handling them within a common representation would greatly simplify motion generation systems. We observe that diverse motion conditions can be naturally formulated as different observation patterns over motion sequences, where each task corresponds to a specific masking strategy. Based on this insight, we introduce \textbf{MotionMaestro}, a unified motion generation framework that learns a shared representation for complete motions and heterogeneous partial observations through masked motion tokenization. MotionMaestro employs a three-stage training strategy that first learns a masked motion tokenizer, then refines its reconstruction ability on clean motions, and finally trains a conditional flow-matching generator in the learned latent space. Furthermore, we introduce an observation map and an observation loss to explicitly preserve provided motion conditions during generation. With this unified representation and conditioning mechanism, MotionMaestro supports text-guided and unconditional synthesis, pose conditioning and partial completion, temporal interpolation, trajectory control, and motion continuation. Experiments on the large-scale RoMo and MotionMillion datasets show state-of-the-art performance across diverse motion generation tasks.
\end{abstract}

\section{Introduction}
\label{sec:intro}

\begin{wrapfigure}{r}{0.5\textwidth}
  \centering
  \vspace{-0.4cm}
  \includegraphics[width=0.5\textwidth]{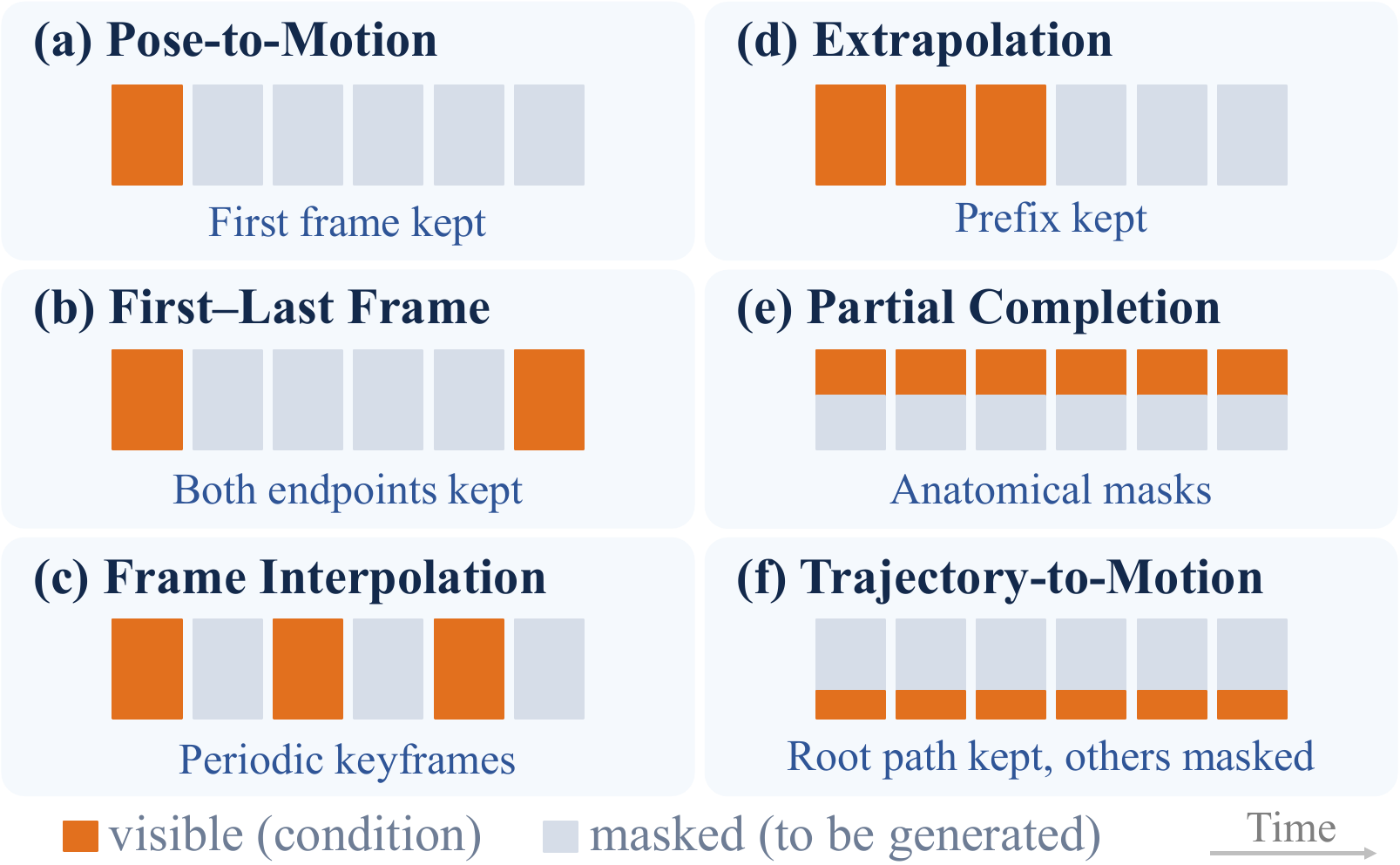}
  \vspace{-0.4cm}
  \caption{\textbf{Motion observation patterns.} Temporal, spatial, and feature-group masks describe the conditions for different motion generation tasks. \textcolor{motionorange}{Orange} marks observed features, while \textcolor{motiongray!60!black}{gray} denotes masked regions to be generated.}
  \label{fig:observationmap}
  \vspace{-0.3cm}
\end{wrapfigure}

Human motion generation aims to synthesize coherent three-dimensional human movements, with applications in animation production and human-centered simulation. A widely studied approach is to guide this process through natural language, with diffusion and token-based models making significant progress in text-to-motion generation~\citep{tevet2022mdm,chen202mld,zhang2023t2mgpt,guo2024momask}. Despite this progress, textual descriptions alone may not provide the fine-grained control required in practical applications. For example, an animator may need to specify key poses, preserve selected body movements, prescribe a trajectory, or extend an existing sequence. To address these control requirements, recent methods have explored keyframe inbetweening~\citep{cohan2024condmdi}, flexible spatial control~\citep{xie2024omnicontrol}, and unified motion generation and editing~\citep{guo2025motionlab,jiang2026motionmaster}. Bringing these capabilities into one framework would allow users to work with different motion conditions without switching between task-specific systems. Meanwhile, recent large-scale datasets, including MotionMillion~\citep{fan2025motionmillion} and RoMo~\citep{zhang2026romo}, provide broader motion coverage and extensive language annotations. Several established controllable generation methods were primarily evaluated on comparatively smaller benchmarks~\citep{xie2024omnicontrol,guo2025motionlab}. The availability of these larger corpora creates an opportunity to study unified generation across both diverse tasks and broader motion distributions.

Alongside advances in datasets and generators, tokenizer research has explored how better latent representations can improve generation. In image generation, MAETok~\citep{chen2025maetok} combines masked learning of semantically informative latents with subsequent decoder-only refinement to improve generation quality and reconstruction fidelity. Inspired by this approach, we observe that \emph{different masking patterns over motion sequences correspond to different motion generation tasks}. Figure~\ref{fig:observationmap} illustrates this connection: retaining sparse frames provides interpolation conditions, retaining selected body parts provides partial completion conditions, and retaining root-motion features provides trajectory conditions. Thus, masking is not merely a representation-learning strategy; in our formulation, \textit{it specifies the task itself by determining which motion features are given and which must be generated}. This motivates training a motion tokenizer with task-aligned masking, preparing its encoder to handle the conditions encountered during generation.

Based on this insight, we propose \textbf{MotionMaestro}, a unified motion generation framework built on \textit{task-aligned masked tokenization}. Our motion tokenizer comprises encoding and decoding components, jointly trained on complete and masked motions with varying temporal coverage, body regions, and feature availability. This training prepares a single encoder for both target representation and condition encoding. Consequently, complete targets and heterogeneous partial conditions are embedded in the same continuous joint--time latent space. The generator thus builds on a representation already trained to handle the incomplete observations. Decoder refinement on clean motions further improves reconstruction fidelity while preserving this shared representation.

To complement these learned representations, we introduce an explicit observation map. 
While condition latents represent motion content, the map identifies which frames, joints, and feature groups are observed. A conditional flow-matching generator~\citep{lipman2022flowmatch} receives both condition latents and the observation map, allowing it to distinguish the available observations from the motion features that remain to be generated. This combination provides a common conditioning interface for tasks with different observation patterns. We further introduce an observation loss to strengthen adherence to the provided conditions. This loss adds direct supervision on observed features in decoded motion sequences. By placing additional emphasis on information that is already known (unmasked motion condition), it encourages the generator to preserve these observations while synthesizing the remaining motion.

Through this shared representation and conditioning interface, a single model supports text-to-motion and unconditional generation, pose conditioning with or without text, first-last frame inbetweening, frame interpolation, motion extrapolation, partial-body completion, trajectory control, and long-sequence generation. We evaluate MotionMaestro on RoMo and MotionMillion against unified baselines. Our results show that a single framework can accommodate diverse motion conditions while achieving high motion generation quality across tasks and datasets, with qualitative examples also suggesting its applicability to text-guided motion editing. Ablation studies further examine the contributions of masked tokenization, the observation map, and the observation loss.

Our contributions are summarized as follows:
\begin{itemize}[leftmargin=2.7em]
\item We present \textbf{MotionMaestro}, a unified framework that supports a broad range of motion generation tasks through one shared representation and conditioning interface.
\item We develop a motion tokenizer with task-aligned masking, enabling a single encoder to embed complete motion targets and heterogeneous partial observations in a shared latent space.
\item We introduce an observation map and a complementary motion-space observation loss to explicitly represent available motion information and encourage consistency with the provided conditions.
\item We demonstrate state-of-the-art performance across diverse tasks on the large-scale RoMo and MotionMillion datasets, significantly outperforming existing unified baselines.
\end{itemize}

\section{Related Work}
\label{sec:related}

\subsection{Human Motion Generation}

\paragraph{Text-to-motion synthesis.}
Text-conditioned motion generation has advanced through generative modeling and representation learning. Diffusion approaches operate directly in motion space~\citep{tevet2022mdm,zhang2024motiondiffuse} or in continuous latents learned by variational autoencoders~\citep{chen202mld}. Discrete methods instead generate quantized motion codes through autoregressive prediction~\citep{zhang2023t2mgpt} or masked modeling with residual-token refinement~\citep{guo2024momask}. Recent work further explores hierarchical flow matching across temporal scales~\citep{li2026motionhiflow}. Alongside these developments, model-scaling studies~\citep{lu2025scamo} and larger motion corpora have supported billion-parameter generators~\citep{fan2025motionmillion,wen2025hy}.

\paragraph{Controllable and unified generation.}
Control beyond text introduces constraints on selected frames, joints, and trajectories. Inpainting and mask-based conditioning support temporal completion and partial-body control~\citep{tevet2022mdm,cohan2024condmdi}. OmniControl~\citep{xie2024omnicontrol} combines analytic spatial guidance with learned realism guidance, while MotionLab~\citep{guo2025motionlab} unifies generation and editing through task instructions and masked generator pretraining. Other unified approaches integrate quantized motion tokens with pretrained language models~\citep{jiang2026motionmaster} or use frame-level operation embeddings to adapt pretrained motion generators~\citep{cong2026umo}. Our MotionMaestro instead pretrains a tokenizer encoder on both complete motions and task-aligned partial observations, establishing a shared latent representation for targets and conditions before generator training. Multimodal unification integrates inputs such as music and speech~\citep{xu2026omnimotionx}, whereas our scope is diverse motion observations with optional text.

\subsection{Motion Tokenization and Masked Representation Learning}

\paragraph{Structured motion latents.}
Beyond the choice between discrete and continuous representations, motion tokenizers differ in how they organize temporal and skeletal information. Some methods encode all joints together into whole-body tokens arranged over time~\citep{zhang2023t2mgpt,guo2024momask}. Others encode body parts separately~\citep{zou2024parco} or represent individual joints in a two-dimensional joint--time token map~\citep{yuan2024mogents}, making spatial structure more explicit. Continuous approaches also incorporate skeletal topologies through skeleton-aware encoding and spatial pooling~\citep{hong2025salad,li2026motionhiflow}. MotionMaestro uses continuous joint--time latents with temporal compression but without joint-axis reduction, retaining explicit joint locations for conditions with different spatial coverage.

\paragraph{Masked tokenizer learning.}
Masked autoencoders learn representations by reconstructing missing inputs~\citep{he2022mae}. MAETok~\citep{chen2025maetok} combines masked image tokenizer learning with auxiliary feature prediction to shape the latent space, followed by decoder-only refinement to improve reconstruction. In motion generation, masking also trains generators through missing-code prediction~\citep{guo2024momask,yuan2024mogents} and masked-motion reconstruction~\citep{guo2025motionlab}. These objectives teach the generator to recover missing motion, rather than prepare the tokenizer encoder to encode incomplete observations. MotionMaestro instead trains its tokenizer on complete motions and task-aligned partial observations. 


\section{MotionMaestro}
\label{sec:method}

\begin{figure*}[t]
\centering
\includegraphics[width=\textwidth]{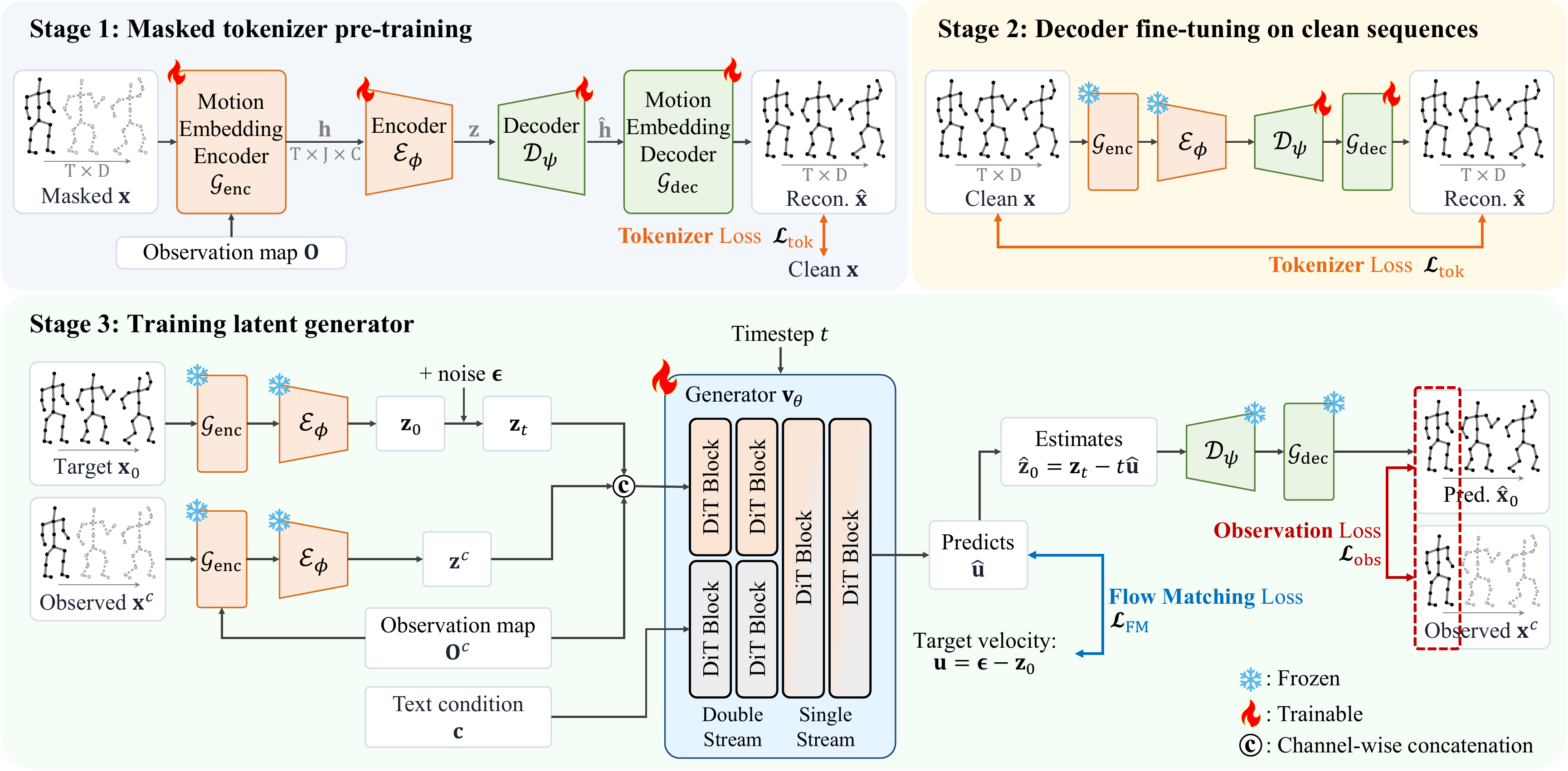}
\caption{\textbf{Overview of MotionMaestro.} Stage~1 trains the tokenizer's encoder and decoder through masked reconstruction. Stage~2 refines only the decoder on clean sequences with the encoder frozen. Stage~3 freezes the entire tokenizer and trains a conditional flow-matching generator. Complete targets and masked conditions share the same encoder, while an observation map identifies available motion information. The observation loss adds motion-space supervision on observed features.}
\label{fig:method}
\end{figure*}

MotionMaestro combines masked motion tokenization with conditional flow matching to support diverse motion generation tasks. Our motion tokenizer adopts an autoencoder architecture with encoding and decoding components. Figure~\ref{fig:method} illustrates the three-stage training procedure of MotionMaestro. Stage~1 jointly trains the tokenizer through masked reconstruction to learn a shared representation of complete motions and partial observations. Stage~2 freezes the encoder and fine-tunes the decoder on clean sequences to improve reconstruction fidelity. Stage~3 freezes the entire tokenizer and trains a conditional flow-matching generator in the learned latent space.

\subsection{Masked Motion Tokenization}
\label{sec:tokenizer}
Let $\mathbf{x}\in\mathbb{R}^{T\times D}$ denote a motion sequence with $T$ frames and $D$ features. The motion representation consists of four feature groups, $\mathcal{G}=\{\texttt{root},\texttt{ric},\texttt{rot},\texttt{vel}\}$, corresponding to root motion, joint positions, joint rotations, and joint velocities, respectively. These features are organized over $J$ nodes comprising a root-motion node and the skeletal joints.

\paragraph{Masking strategy.}
We describe each motion condition with a binary mask tensor as an observation map $\mathbf{O}\in\{0,1\}^{T\times J\times G}$, where $G$ is the number of feature groups with $G=|\mathcal{G}|=4$. Each entry indicates whether a defined feature group at a given frame and node is observed ($1$) or missing ($0$) (masked). The motion embedding encoder $\mathcal{G}_{\mathrm{enc}}$ uses this map to construct joint-time embeddings $\mathbf{h}$, while the motion embedding decoder $\mathcal{G}_{\mathrm{dec}}$ reconstructs motion features from the decoder output $\hat{\mathbf{h}}$:
\begin{equation}
\mathbf{h}=\mathcal{G}_{\mathrm{enc}}(\mathbf{x},\mathbf{O}), \qquad \hat{\mathbf{x}}=\mathcal{G}_{\mathrm{dec}}(\hat{\mathbf{h}}),
\label{eq:motion_group_mapping}
\end{equation}
where $\mathbf{h},\hat{\mathbf{h}}\in\mathbb{R}^{T\times J\times C}$. For each group $g\in\mathcal{G}$, we arrange its features as $\mathbf{x}_g\in\mathbb{R}^{T\times J_g\times C_g}$ and project them to the same channel dimension $C$ using a pointwise projection $\Phi_g$ shared across its $J_g$ nodes. We retain observed projections and replace missing ones with a learnable embedding $\mathbf{e}_{\mathrm{mask}}\in\mathbb{R}^{C}$ shared across frames, nodes, and groups. Adding a group identity embedding $\mathbf{e}_g$ gives group-level embeddings $\mathbf{h}_g$:
\begin{equation}
\mathbf{h}_g=\mathbf{O}_g\odot\Phi_g(\mathbf{x}_g)+(1-\mathbf{O}_g)\odot\mathbf{e}_{\mathrm{mask}}+\mathbf{e}_g,
\label{eq:feature_mask}
\end{equation}
where $\odot$ denotes element-wise multiplication and $\mathbf{O}_g$ is the observation map for group $g$, broadcast across channels. At each frame and node, we channel-wise 
concatenate the embeddings $\mathbf{h}_g$ of the feature groups defined for that node. We project the concatenated vector to $C$ channels and add node and temporal positional embeddings to obtain $\mathbf{h}$.

\begin{wrapfigure}{r}{0.52\textwidth}
  \centering
  \vspace{-0.4cm}
  \includegraphics[width=0.52\textwidth]{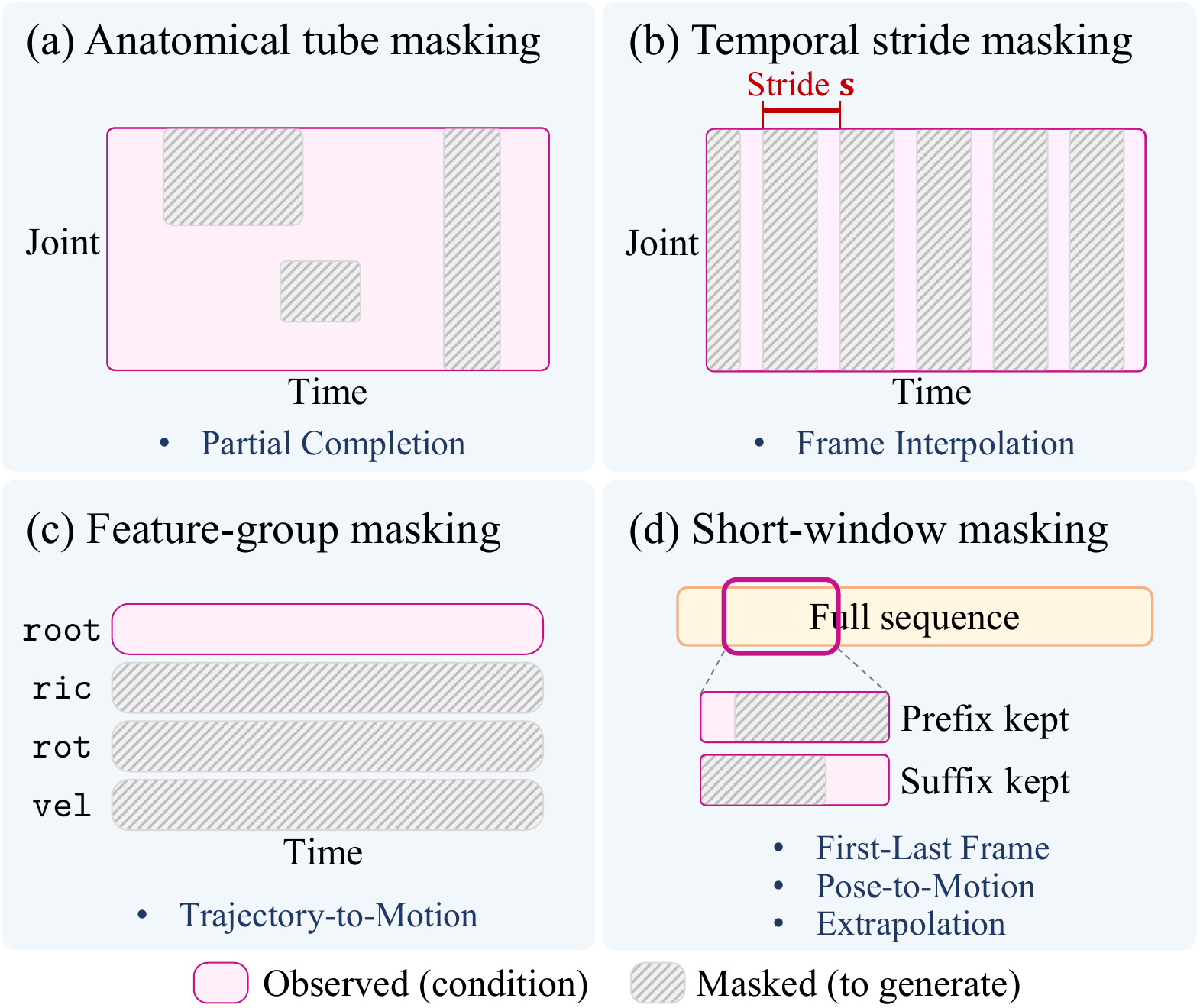}
  \vspace{-0.6cm}
  \caption{\textbf{Task-aligned masking for motion tokenization.} Anatomical tube, temporal stride, feature-group, and short-window masking expose the tokenizer to the motion conditions associated with different tasks. \textcolor{motionpink}{Pink} regions denote observed inputs, while hatched regions are masked.}
  \vspace{-0.3cm}
  \label{fig:masking}
\end{wrapfigure}
Figure~\ref{fig:masking} illustrates the four task-aligned strategies used in our structured masking scheme. (a) \emph{Anatomical tube masking} masks body regions over temporal intervals, respecting skeletal connectivity when selecting joints to model partial-body completion. (b) \emph{Temporal stride masking} retains uniformly spaced frames and masks the intervening frames to model frame interpolation. (c) \emph{Feature-group masking} randomly removes selected feature groups; for example, retaining root-motion features while masking the other groups provides trajectory conditions. (d) \emph{Short-window masking} crops a short motion subsequence and retains an observed prefix or suffix within it. A few observed frames do not uniquely determine a long motion sequence, making full-sequence reconstruction an overly ambiguous objective for the tokenizer. We therefore restrict reconstruction to the cropped window, allowing the encoder to learn representations for isolated poses, endpoint frames, and short prefixes.

\paragraph{Tokenizer training.}
The graph-temporal encoder $\mathcal{E}_{\phi}$ and decoder $\mathcal{D}_{\psi}$ adapt the FLUX.2 autoencoder architecture~\citep{flux-2-2025}, replacing image-grid convolutions with temporal convolutions and skeletal graph aggregation. The encoder reduces the temporal resolution by a factor of four while preserving the $J$ nodes, producing $\mathbf{z}\in\mathbb{R}^{K\times J\times C_z}$, where $K=\widetilde{T}/4$ and $\widetilde{T}$ is the padded sequence length. In Stage~1, we jointly train the tokenizer, including all components $\mathcal{G}_{\mathrm{enc}}$, $\mathcal{E}_{\phi}$, $\mathcal{D}_{\psi}$ and $\mathcal{G}_{\mathrm{dec}}$, on masked and clean motion sequences through reconstruction:
\begin{equation}
\hat{\mathbf{x}}=\mathcal{G}_{\mathrm{dec}}\!\left(\mathcal{D}_{\psi}\!\left(\mathcal{E}_{\phi}\!\left(\mathcal{G}_{\mathrm{enc}}(\mathbf{x},\mathbf{O})\right)\right)\right).
\label{eq:tokenizer_reconstruction}
\end{equation}
For inputs with full-motion reconstruction targets, we minimize Tokenizer Loss $\mathcal{L}_{\mathrm{tok}}$ as
\begin{equation}
\mathcal{L}_{\mathrm{tok}}=\mathbb{E}_{\mathbf{x},\mathbf{O}}\!\left[\mathcal{L}_{\mathrm{rec}}+\lambda_p\mathcal{L}_{\mathrm{path}}+\lambda_y\mathcal{L}_{\mathrm{yaw}}\right],
\label{eq:tokenizer_loss}
\end{equation}
where all three terms use an $\ell_1$ loss between the reconstruction and the target. $\mathcal{L}_{\mathrm{rec}}$ averages normalized feature reconstruction errors across groups, while $\mathcal{L}_{\mathrm{path}}$ and $\mathcal{L}_{\mathrm{yaw}}$ supervise the recovered root trajectory and heading, respectively. We set $\lambda_p=\lambda_y=0.1$. For trajectory-only inputs, supervision is restricted to the observed trajectory features and recoverable root quantities. All losses exclude padded frames and structurally undefined features.

In Stage~1, the decoder reconstructs motions from latent representations of both complete and incomplete observations. At inference, however, it receives generated latents intended to represent complete motions. Therefore, in Stage~2, we freeze the encoder components $\mathcal{G}_{\mathrm{enc}}$ and $\mathcal{E}_{\phi}$ and fine-tune only the decoder components $\mathcal{D}_{\psi}$ and $\mathcal{G}_{\mathrm{dec}}$. We optimize $\mathcal{L}_{\mathrm{tok}}$ using clean sequences, with all defined features marked as observed. This refines reconstruction from clean-motion latents while preserving the learned conditioning representation.


\subsection{Unified Conditional Generation}
\label{sec:generator}

\paragraph{Unified conditioning.}
In Stage~3, we freeze the entire tokenizer and use the same encoder components for clean targets and motion conditions. We denote the target motion sequence by $\mathbf{x}_0$ and its partial observation by $\mathbf{x}^{c}$:
\begin{equation}
\mathbf{z}_0=\mathcal{E}_{\phi}\!\left(\mathcal{G}_{\mathrm{enc}}(\mathbf{x}_0,\mathbf{1})\right), \qquad \mathbf{z}^{\mathrm{enc}}=\mathcal{E}_{\phi}\!\left(\mathcal{G}_{\mathrm{enc}}(\mathbf{x}^{c},\mathbf{O}^{c})\right),
\label{eq:motion_latents}
\end{equation}
where $\mathbf{1}$ marks all defined feature groups as observed and $\mathbf{O}^{c}$ specifies the supplied motion observations.
For short-window conditions, we construct $\mathbf{z}^{c}$ by placing $\mathbf{z}^{\mathrm{enc}}$ at the corresponding locations on the target latent grid and filling the remaining locations with a learnable embedding $\mathbf{e}_{\mathrm{mask}}^{z}\in\mathbb{R}^{C_z}$.
Otherwise, we use $\mathbf{z}^{c}=\mathbf{z}^{\mathrm{enc}}$. The observation map $\mathbf{O}^{c}$ is further reshaped into a latent-space representation by folding every four frames into channels, obtaining latent observation map $\mathbf{m}\in\{0,1\}^{K\times J\times 4G}$. Finally, at flow time $t$, the generator receives the channel-wise concatenation of the noisy target latent $\mathbf z_t$, condition latent $\mathbf z^c$, and latent observation map $\mathbf m$.

\paragraph{Generator training.}
We train the generator from scratch using FLUX.2-style transformer blocks~\citep{flux-2-2025}, with text features $\mathbf{c}$ provided by a frozen text encoder~\citep{chung2024scaling}. Double-stream blocks process motion and text with separate parameters while exchanging information through joint attention. Single-stream blocks then process their combined sequence, with only motion tokens passed to the velocity head. Following flow matching~\citep{lipman2022flowmatch}, we sample noise $\boldsymbol{\epsilon}\sim\mathcal{N}(\mathbf{0},\mathbf{I})$ and $t=\operatorname{sigmoid}(\eta)$ with $\eta\sim\mathcal{N}(0,1)$, and define
\begin{equation}
\mathbf{z}_t=(1-t)\mathbf{z}_0+t\boldsymbol{\epsilon}, \qquad \mathbf{u}=\frac{\partial\mathbf{z}_t}{\partial t}=\boldsymbol{\epsilon}-\mathbf{z}_0.
\label{eq:flow_path}
\end{equation}

The generator predicts the conditional velocity field $\mathbf{v}_{\theta}$, from which we estimate the clean latent as:
\begin{equation}
\hat{\mathbf{z}}_0^{\theta}=\mathbf{z}_t-t\,\mathbf{v}_{\theta}(\mathbf{z}_t,t;\mathbf{z}^{c},\mathbf{m},\mathbf{c}),
\label{eq:clean_latent_estimate}
\end{equation}
and decode it into $\hat{\mathbf{x}}_0^{\theta}=\mathcal{G}_{\mathrm{dec}}\!\left(\mathcal{D}_{\psi}(\hat{\mathbf{z}}_0^{\theta})\right)$. The Flow Matching Loss $\mathcal{L}_{\mathrm{FM}}$ is used to supervise the predicted velocity. We additionally introduce Observation Loss $\mathcal{L}_{\mathrm{obs}}$ to provide motion-space supervision, encouraging the generated motion to remain consistent with the supplied observations:
\begin{equation}
\mathcal{L}_{\mathrm{FM}}=\mathbb{E}\!\left[\left\|\mathbf{v}_{\theta}(\mathbf{z}_t,t;\mathbf{z}^{c},\mathbf{m},\mathbf{c})-\mathbf{u}\right\|_2^2\right], \qquad \mathcal{L}_{\mathrm{obs}}=\mathbb{E}\!\left[\ell_1(\hat{\mathbf{x}}_0^{\theta},\mathbf{x}_0;\mathbf{O}^{c})\right],
\label{eq:flow_loss}
\end{equation}
where $\ell_1(\cdot,\cdot;\mathbf{O}^{c})$ computes reconstruction errors only over features marked as observed, excluding padded frames. The full generator objective is
\begin{equation}
\mathcal{L}_{\mathrm{gen}}=\mathcal{L}_{\mathrm{FM}}+\lambda_o\mathcal{L}_{\mathrm{obs}},
\label{eq:generator_loss}
\end{equation}
where $\lambda_o=0.1$. Decoder parameters remain frozen, while gradients from $\mathcal{L}_{\mathrm{obs}}$ propagate through the decoder to the generator. 
Training includes examples without motion conditioning, and we randomly drop text conditioning to support text-to-motion, unconditional generation, and text classifier-free guidance.

\paragraph{Inference.}
We initialize $\mathbf{z}_1$ with Gaussian noise and integrate the predicted velocity field from $t=1$ to $t=0$ using Euler steps, optionally applying text classifier-free guidance while retaining the motion conditions. The frozen decoder maps the resulting latent to the generated motion sequence. All supported tasks use the same interface for motion and text conditioning.


\section{Experiments}
\label{sec:experiments}

\subsection{Datasets and evaluation protocol}
\label{sec:dataset}

\paragraph{MotionMillion dataset.}
MotionMillion~\citep{fan2025motionmillion} combines motions
reconstructed from web videos with existing motion datasets,
comprising approximately two million sequences spanning
over 2,000 hours.
Its captions describe motion content, style, and context,
with each original description augmented by 20 semantically
consistent paraphrases.

\paragraph{RoMo dataset.}
RoMo~\citep{zhang2026romo} contains approximately 814K motion clips spanning 1,238 hours, with filtering
to remove static and artifact-prone sequences.
Each clip includes five caption variants and hierarchical
annotations at the category, subcategory, and atomic-action
levels.

\begin{table*}[t]
\centering

\caption{\textbf{Unified motion generation on RoMo.} Each method is evaluated on its supported tasks.
FI and partial completion report MPJPE (mm) over unobserved frames and joints, respectively; the remaining tasks report FID. Lower is better. \textbf{Bold} and \underline{underlined} values mark the best and second-best results per task. Results for MotionMaestro-5B, scaled up from our default 1.3B model, are provided in the \textit{Appendix}.}

\label{tab:main_romo}
\scriptsize
\setlength{\tabcolsep}{3pt}
\renewcommand{\arraystretch}{1.1}
\resizebox{\textwidth}{!}{%
\begin{tabular}{lccccccccccc}
\toprule
\textbf{Method} & \textbf{Venue}
& \makecell{\textbf{Uncond.}\\FID$\downarrow$} & \makecell{\textbf{T2M}\\FID$\downarrow$} & \makecell{\textbf{P2M}\\FID$\downarrow$} & \makecell{\textbf{TP2M}\\FID$\downarrow$} & \makecell{\textbf{FLF}\\FID$\downarrow$} & \makecell{\textbf{FI}\\MPJPE$\downarrow$} & \makecell{\textbf{Extrap.}\\FID$\downarrow$} & \makecell{\textbf{Partial}\\MPJPE$\downarrow$} & \makecell{\textbf{Trajectory}\\FID$\downarrow$} & \makecell{\textbf{Long}\\FID$\downarrow$} \\
\midrule
MDM~\citep{tevet2022mdm} & ICLR2023
& 507.6 & 450.7 & 518.4 & 456.1 & 523.3 & 117.6 & 342.8 & 278.9 & -- & 510.9 \\
OmniControl~\citep{xie2024omnicontrol} & ICLR2024
& 429.1 & 199.4 & 771.0 & 312.8 & 753.2 & 875.9 & 740.3 & 627.3 & 283.7 & 398.6 \\
MotionLab~\citep{guo2025motionlab} & ICCV2025
& \underline{76.7} & \underline{101.5} & \underline{53.5} & \underline{46.5} & \underline{30.0} & \underline{12.7} & \underline{28.1} & \underline{71.8} & \underline{89.3} & \underline{123.5} \\
\rowcolor{orange!12}
\textbf{MotionMaestro (Ours)} & --
& \textbf{20.4} & \textbf{14.9} & \textbf{34.8} & \textbf{28.7} & \textbf{20.8} & \textbf{6.7} & \textbf{19.4} & \textbf{61.0} & \textbf{59.2} & \textbf{26.2} \\
\bottomrule
\end{tabular}%
}
\end{table*}

\begin{table*}[t]
\centering
\caption{
\textbf{Unified motion generation and computational cost
on MotionMillion.}
Evaluation settings, metrics, and notation follow
Table~\ref{tab:main_romo}.
We additionally report computational cost.
}
\label{tab:main_mm}
\scriptsize
\setlength{\tabcolsep}{3pt}
\renewcommand{\arraystretch}{1.1}
\resizebox{\textwidth}{!}{%
\begin{tabular}{lccccccccccccc}
\toprule
\textbf{Method}
& \makecell{\textbf{Uncond.}\\FID$\downarrow$} & \makecell{\textbf{T2M}\\FID$\downarrow$} & \makecell{\textbf{P2M}\\FID$\downarrow$} & \makecell{\textbf{TP2M}\\FID$\downarrow$} & \makecell{\textbf{FLF}\\FID$\downarrow$} & \makecell{\textbf{FI}\\MPJPE$\downarrow$} & \makecell{\textbf{Extrap.}\\FID$\downarrow$} & \makecell{\textbf{Partial}\\MPJPE$\downarrow$} & \makecell{\textbf{Trajectory}\\FID$\downarrow$}& \makecell{\textbf{Long}\\FID$\downarrow$}
& \makecell{\textbf{Inference}\\Time (s)}
& \makecell{\textbf{Memory}\\GB}
& \makecell{\textbf{Params}\\Billion (B)}\\
\midrule
MDM & 427.7 & 297.6 & 523.8 & 334.4 & 530.6 & 122.8 & 370.4 & 206.8 & -- & 336.2 & 4.69 & 0.4 & 0.081 \\
OmniControl & 376.3 & 153.9 & 366.8 & 193.8 & 365.5 & 375.4 & 551.3 & 349.6 & 276.6 & 177.9 & 61.58 & 0.5 & 0.102 \\
MotionLab & \underline{135.1} & \underline{61.4} & \underline{74.4} & \textbf{49.4} & \underline{50.7} & \underline{12.3} & \underline{38.8} & \underline{71.1} & \textbf{94.4} & \underline{108.6} & 0.97 & 2.6 & 0.367\\
\rowcolor{orange!12}
\textbf{MotionMaestro} & \textbf{67.2} & \textbf{36.2} & \textbf{74.3} & \underline{52.8} & \textbf{44.5} & \textbf{3.8} & \textbf{36.9} & \textbf{61.3} & \underline{118.2} & \textbf{46.4} & 1.28 & 11.7 & 1.285 \\
\bottomrule
\end{tabular}%
}
\end{table*}

\paragraph{Evaluation protocol.}
We evaluate one generator checkpoint per dataset across
ten tasks: unconditional generation, text-to-motion (T2M),
pose-to-motion (P2M), text+pose-to-motion
(TP2M), first-last frame (FLF),
frame interpolation (FI), extrapolation, partial completion, trajectory-to-motion, and long sequence generation.
For reconstruction-focused tasks (FI and partial completion), we assess reconstruction accuracy using mean per-joint position error (MPJPE, in mm), computed over unobserved frames and joints, respectively. For the remaining generation-focused tasks, we use Fr\'echet Inception Distance (FID) to assess distributional quality by comparing the feature distributions of generated and ground-truth motions. For both metrics, lower is better.

\subsection{Implementation and comparisons}
\label{sec:exp_details}

\paragraph{Implementation details.}
Our tokenizer produces 16-channel latents over 23 joint slots
at temporal stride~4.
Our generator uses 32 transformer blocks with hidden width
1,536 and 24 attention heads, conditioned on features
from a frozen FLAN-T5-XL text encoder.
We train on two NVIDIA B200 GPUs using AdamW and
bfloat16 mixed precision.
Stages~1--3 run for 100k, 20k, and 200k updates with
peak learning rates of $10^{-4}$, $2\times10^{-5}$,
and $10^{-4}$, respectively, using cosine decay.
Generator training takes approximately 48 hours per dataset.
Inference uses EMA weights with decay $0.999$ and
50 Euler steps, requiring 50 function evaluations
without classifier-free guidance and 100 with text guidance.
Further architecture and training details are provided
in the \textit{Appendix}.

\paragraph{Baselines.}
We compare against the motion-space diffusion model MDM~\citep{tevet2022mdm}, the spatially controllable diffusion model OmniControl~\citep{xie2024omnicontrol}, and the unified flow-based model MotionLab~\citep{guo2025motionlab}. All baselines are retrained on each dataset using the same data splits and evaluated on their supported tasks under the same test conditions and metrics.

\begin{figure*}[t]
\centering
\includegraphics[width=\textwidth]{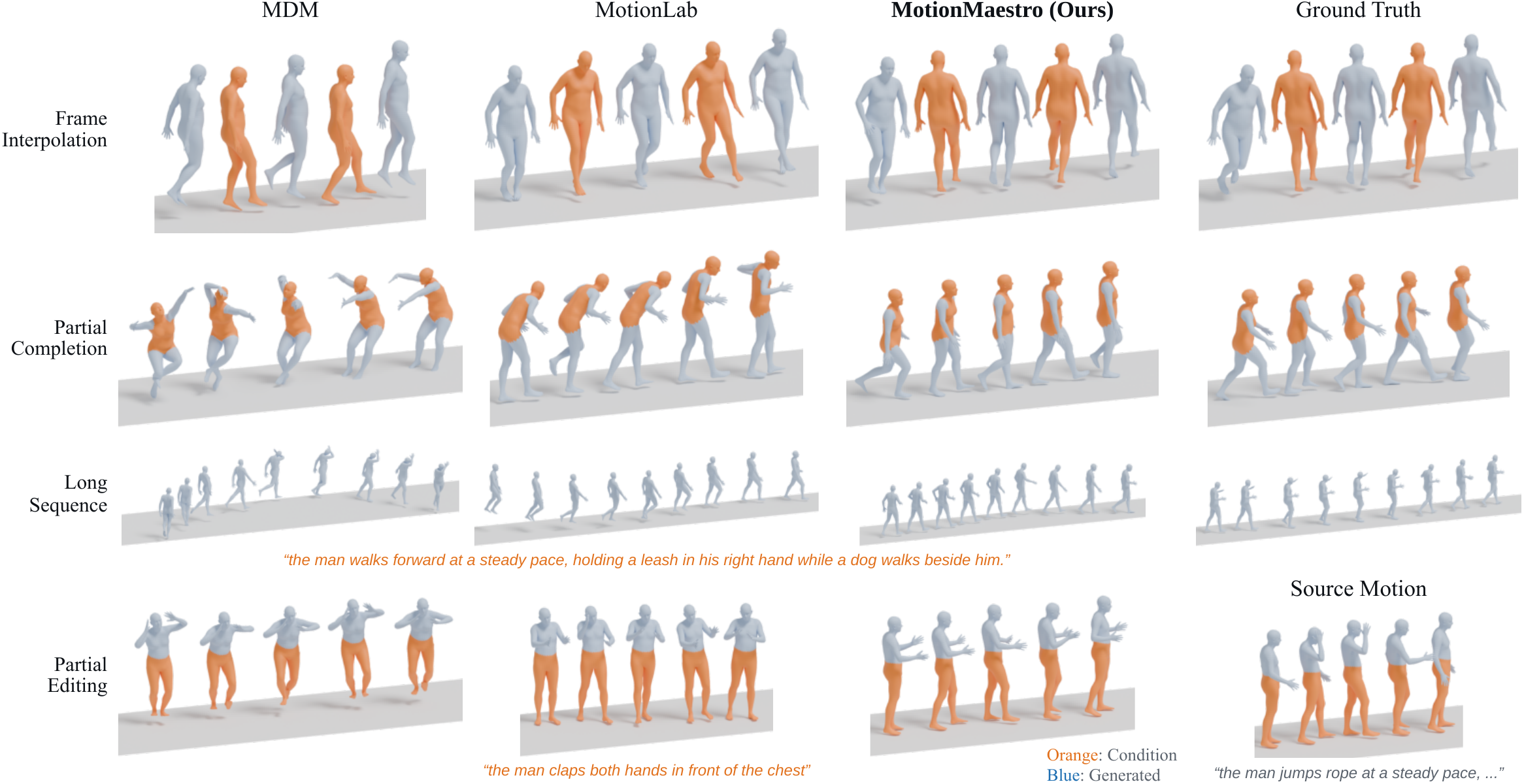}
\caption{
\textbf{Qualitative comparison and zero-shot motion editing.}
We compare MotionMaestro with MDM and MotionLab on
frame interpolation, partial completion,
long sequence generation, and text-guided partial editing. 
The partial editing example demonstrates MotionMaestro's editing capability without editing-specific training or fine-tuning. See the \textit{Appendix} for additional results.}
\label{fig:visual}
\end{figure*}

\begin{table}[t]
\centering
\begin{minipage}{0.55\textwidth}
\centering
\caption{\textbf{Frame interpolation with varying observation stride on RoMo.}}
\vspace{-0.2cm}
\label{tab:fi_stride}
\scriptsize
\setlength{\tabcolsep}{4pt}
\renewcommand{\arraystretch}{1.1}
\resizebox{\textwidth}{!}{%
\begin{tabular}{lcccccccc}
\toprule
\multirow{2}{*}{\textbf{Method}} & \multicolumn{2}{c}{$s{=}2$} & \multicolumn{2}{c}{$s{=}8$} & \multicolumn{2}{c}{$s{=}16$} & \multicolumn{2}{c}{$s{=}32$} \\
\cmidrule(lr){2-9}
 & MPJPE$\downarrow$ & FID$\downarrow$ & MPJPE$\downarrow$ & FID$\downarrow$ & MPJPE$\downarrow$ & FID$\downarrow$ & MPJPE$\downarrow$ & FID$\downarrow$ \\
\midrule
MDM & 59.3 & 61.9 & 252.2 & 282.5 & 375.2 & 403.6 & 431.3 & 464.0 \\
OmniControl & 803.2 & 747.8 & 929.2 & 782.7 & 943.3 & 776.1 & 930.6 & 745.8 \\
MotionLab & \underline{11.1} & \underline{10.7} & \underline{18.9} & \underline{14.1} & \underline{35.4} & \underline{21.4} & \underline{57.5} & \underline{30.4} \\
\rowcolor{orange!12}
\textbf{MotionMaestro} & \textbf{4.8} & \textbf{3.9} & \textbf{12.7} & \textbf{4.7} & \textbf{30.1} & \textbf{12.9} & \textbf{54.1} & \textbf{18.6} \\
\bottomrule
\end{tabular}}
\end{minipage}
\hfill
\begin{minipage}{0.42\textwidth}
\centering
\caption{\textbf{Partial completion under different observed body parts on RoMo.}}
\vspace{-0.2cm}
\label{tab:partial}
\scriptsize
\setlength{\tabcolsep}{3pt}
\renewcommand{\arraystretch}{1.1}
\resizebox{\textwidth}{!}{%
\begin{tabular}{lcccccc}
\toprule
\multirow{2}{*}{\textbf{Method}} & \multicolumn{2}{c}{torso} & \multicolumn{2}{c}{head$+$hands} & \multicolumn{2}{c}{one arm} \\
\cmidrule(lr){2-7}
& MPJPE$\downarrow$ & FID$\downarrow$ & MPJPE$\downarrow$ & FID$\downarrow$ & MPJPE$\downarrow$ & FID$\downarrow$ \\
\midrule
MDM & \underline{301.4} & 553.8 & 346.0 & 516.8 & 313.5 & 561.4 \\
OmniControl & 535.3 & 477.5 & 627.3 & 631.5 & 297.4 & 293.1 \\
MotionLab & 378.1 & \underline{379.3} & \underline{73.8} & \underline{35.3} & \underline{81.2} & \textbf{32.7} \\
\rowcolor{orange!12}
\textbf{MotionMaestro} & \textbf{109.3} & \textbf{44.7} & \textbf{68.5} & \textbf{31.8} & \textbf{79.5} & \underline{39.0} \\
\bottomrule
\end{tabular}}
\end{minipage}
\hfill
\end{table}

\subsection{Main Evaluation}
\label{sec:results}

\paragraph{Quantitative comparison.}
Tables~\ref{tab:main_romo} and~\ref{tab:main_mm} compare MotionMaestro with existing methods on unified motion generation across ten tasks. FI uses an observation stride of $s=4$, and the partial completion task evaluates completion conditioned on upper-body joint positions.
MotionMaestro achieves the best results on all ten tasks on RoMo and eight of ten on MotionMillion. For text-to-motion on RoMo, MotionMaestro achieves an FID of 14.9, outperforming MotionLab's 101.5 by 86.6. The gains also extend to long sequence generation on MotionMillion: MotionMaestro achieves an FID of 46.4, improving over MotionLab's 108.6 by 62.2.

\paragraph{Qualitative comparison.}
Figure~\ref{fig:teaser} shows generation under diverse textual, spatial, and temporal conditions
with a single model.
Figure~\ref{fig:visual} compares MotionMaestro with MDM and MotionLab. MotionMaestro better preserves observed keyframes and reconstructs intermediate poses close to the ground truth in frame interpolation. For partial completion, MotionMaestro retains observed regions while generating realistic movements for missing joints. These movements can differ from the reference while
remaining compatible with the observations, reflecting
the multiple plausible solutions admitted by partial conditions.

\paragraph{Frame interpolation.} Table~\ref{tab:fi_stride} evaluates frame interpolation across different observation densities. MotionMaestro achieves the lowest (best) MPJPE and FID at every evaluated stride. As observations become sparser, accurate recovery of the reference poses becomes more challenging, resulting in higher reconstruction error. Nevertheless, at the sparsest evaluated stride, $s=32$, MotionMaestro achieves an FID of 18.6, substantially lower than MotionLab's 30.4. These results indicate that MotionMaestro retains its advantage in generation quality even with sparse keyframes.

\paragraph{Partial completion.} 
Table~\ref{tab:partial} examines how partial completion performance varies with different observed body regions. MotionMaestro achieves the best MPJPE across all evaluated masks, with the strongest improvement under torso conditioning: MPJPE decreases from 378.1 to 109.3\,mm relative to MotionLab, accompanied by a substantial improvement in FID from 379.3 to 44.7. Both metrics also improve under head$+$hands conditioning. Under one-arm conditioning, MotionMaestro achieves higher FID but lower MPJPE than MotionLab.

\paragraph{Zero-shot motion editing.} Without editing-specific training or fine-tuning, MotionMaestro can combine source-motion observations with a new text prompt for partial editing. Figure~\ref{fig:visual} shows the model replacing rope-jumping arm movements with clapping while retaining the source lower-body motion. This qualitative example provides preliminary evidence that the unified conditioning interface can extend to motion editing; systematic quantitative evaluation remains future work.

\paragraph{Computational cost.} Table~\ref{tab:main_mm} reports the computational requirements under the evaluated inference settings. MotionMaestro takes 1.28\,sec per generation compared with 0.97\,sec for MotionLab, while remaining faster than MDM and OmniControl. Memory usage is 11.7\,GB for MotionMaestro and 2.6\,GB for MotionLab. Recent studies have used large-scale motion datasets to train billion-parameter generators~\citep{fan2025motionmillion,wen2025hy}. Our model scale is consistent with this trend toward scaling both motion data and model capacity.

\begin{table*}[t]
\centering
\caption{\textbf{Ablations of tokenizer training and observation conditioning on RoMo.} The upper block compares training on clean sequences only, random masking, and structured masking (MotionMaestro, bottom row). The lower block uses the same tokenizer trained with structured masking and separately removes the generator's observation map input $\mathbf{O}^{c}$ or observation loss $\mathcal{L}_{\mathrm{obs}}$.}

\label{tab:ablation_masking_main}
\scriptsize
\setlength{\tabcolsep}{3pt}
\renewcommand{\arraystretch}{1.1}
\resizebox{0.8\textwidth}{!}{%
\begin{tabular}{lcccccccccc}
\toprule
\textbf{Method}
& \makecell{\textbf{Uncond.}\\FID$\downarrow$} & \makecell{\textbf{T2M}\\FID$\downarrow$} & \makecell{\textbf{P2M}\\FID$\downarrow$} & \makecell{\textbf{TP2M}\\FID$\downarrow$} & \makecell{\textbf{FLF}\\FID$\downarrow$} & \makecell{\textbf{FI}\\MPJPE$\downarrow$} & \makecell{\textbf{Extrap.}\\FID$\downarrow$} & \makecell{\textbf{Partial}\\MPJPE$\downarrow$} & \makecell{\textbf{Trajectory}\\FID$\downarrow$} & \makecell{\textbf{Long}\\FID$\downarrow$} \\
\midrule
\rowcolor{black!5}
\multicolumn{11}{l}{\textit{Tokenizer training (Stage~1)}} \\
Clean sequences only
& 35.5 & 18.3 & 38.3 & 31.0 & 24.2 & 7.8 & 22.1 & 61.8 & 72.9 & 41.2 \\
Random masking
& 22.3 & 15.6 & 37.8 & 30.6 & 21.0 & 7.5 & 21.3 & 63.7 & 59.7 & 36.2 \\
\midrule
\rowcolor{black!5}
\multicolumn{11}{l}{\textit{Generator training (Stage~3)}} \\
w/o observation map $\mathbf{O}^c$
& \textbf{18.6} & \underline{15.0} & 36.9 & \underline{29.6} & 20.9 & \textbf{6.2} & \underline{19.6} & 61.3 & \textbf{57.5} & 30.9 \\
w/o $\mathcal{L}_{\mathrm{obs}}$
& 21.8 & 15.2 & \underline{35.9} & 29.8 & \textbf{20.1} & 6.8 & 19.9 & \underline{61.1} & \underline{58.9} & \underline{28.7} \\
\midrule
\rowcolor{orange!12}
\textbf{MotionMaestro (Ours)}
& \underline{20.4} & \textbf{14.9} & \textbf{34.8} & \textbf{28.7} & \underline{20.8} & \underline{6.7} & \textbf{19.4} & \textbf{61.0} & 59.2 & \textbf{26.2} \\
\bottomrule
\end{tabular}}
\end{table*}

\subsection{Ablations}
\label{sec:ablations}

\paragraph{Masking strategy.} Table~\ref{tab:ablation_masking_main} compares three tokenizer training strategies: clean sequences only, random masking, and structured masking (MotionMaestro). All three training strategies use the same tokenizer architecture, decoder refinement, and generator training protocol, with matched observation ratios for random and structured masking. Structured masking outperforms clean sequences only and random masking across all evaluated tasks, including unconditional generation and text-to-motion, where no motion observations are provided at inference. For long sequence generation, structured masking improves FID to 26.2, compared to 41.2 for clean sequences only and 36.2 for random masking. These results demonstrate the effectiveness of task-aligned masking in tokenizer training for unified motion generation.

\paragraph{Observation conditioning.} Table~\ref{tab:ablation_masking_main} evaluates the individual contributions of the observation map $\mathbf{O}^{c}$ and observation loss $\mathcal{L}_{\mathrm{obs}}$ by removing each separately. For a controlled comparison, all three settings use the same tokenizer trained with structured masking and the same motion-condition latents. Using both components improves performance in most evaluated settings compared with removing either component. The shared benefit appears in pose-to-motion generation, where the full model improves FID to 34.8 from 36.9 (w/o observation map) and 35.9 (w/o $\mathcal{L}_{\mathrm{obs}}$). These results highlight the benefits of explicitly identifying observed motion features and supervising their reconstruction, alongside conditioning on the learned motion latents. Additional evaluation of how accurately the generated motions preserve the supplied observations is provided in the \textit{Appendix}.

\section{Conclusion}
\label{sec:conclusion}

We presented \textbf{MotionMaestro}, a unified motion generation framework that connects masked motion tokenization with conditional flow matching. Task-aligned masking prepares a shared encoder to represent complete motions and partial conditions in the same latent space. An observation map explicitly identifies available motion information, while an observation loss encourages consistency with the provided conditions. Qualitative and quantitative results on the large-scale RoMo and MotionMillion datasets show state-of-the-art performance across diverse tasks, further illustrating the framework's potential for motion editing.

\clearpage
\appendix

\section*{Appendix}
\noindent
This \textit{Appendix} provides additional evidence
and implementation details for MotionMaestro.
Section~\ref{sec:supp_additional_experiments} examines tokenizer
reconstruction, decoder refinement, observation preservation,
and sampling variability.
Section~\ref{sec:supp_additional_results} compares the default and
5B variants, evaluates frame interpolation and partial completion
under different observations, and analyzes physical plausibility
and text guidance.
Sections~\ref{app:mm_data}, \ref{app:mm_implementation},
and~\ref{app:mm_protocol} describe data and preprocessing,
architecture and training, and evaluation protocols and baseline
adaptations, respectively.
Unless otherwise specified, MotionMaestro denotes the default
1.3B model; MotionMaestro-5B denotes the 5B variant.

\begin{table}[h]
\centering
\caption{Overview of the \textit{Appendix}.}
\label{tab:supple_overview}
\scriptsize
\setlength{\tabcolsep}{6pt}
\renewcommand{\arraystretch}{1.1}
\resizebox{0.55\textwidth}{!}{%
\begin{tabular}{cl}
\toprule
\rowcolor{orange!12}
\textbf{Section} & \textbf{Contents} \\
\midrule
Section~\ref{sec:supp_additional_experiments} & Additional ablations and discussion \\
Section~\ref{sec:supp_additional_results} & Additional experiments \\
Section~\ref{app:mm_data} & Data and preprocessing \\
Section~\ref{app:mm_implementation} & Architecture and training details \\
Section~\ref{app:mm_protocol} & Evaluation protocol and baseline adaptations \\
\bottomrule
\end{tabular}%
}
\end{table}

\section{Additional Ablations and Discussion}
\label{sec:supp_additional_experiments}
\label{app:mm_results}

We first examine whether the tokenizer supports both complete-motion
representation and partial-condition encoding, then assess the
effects of decoder refinement and observation-aware generation.
Repeated sampling distinguishes checkpoint performance from
variation due to sampling noise.

\begin{table*}[t]
\centering
\caption{
\textbf{Clean-motion reconstruction after Stage~1 on RoMo.}
We evaluate complete-motion reconstruction from unmasked
inputs before decoder refinement.
Lower is better for all metrics.
\textbf{Bold} and \underline{underlined} values denote
the best and second-best results in each column, respectively.
}
\label{tab:ae_clean_stage1}
\scriptsize
\setlength{\tabcolsep}{5pt}
\renewcommand{\arraystretch}{1.1}
\resizebox{0.8\textwidth}{!}{%
\begin{tabular}{lcccccc}
\toprule
\textbf{Tokenizer}
& \makecell{\textbf{MPJPE}\\local (mm)$\downarrow$}
& \makecell{\textbf{MPJPE}\\world (mm)$\downarrow$}
& \makecell{\textbf{Rotation}\\(deg)$\downarrow$}
& \makecell{\textbf{Root path}\\(mm)$\downarrow$}
& \makecell{\textbf{Final pos.}\\(mm)$\downarrow$}
& \makecell{\textbf{Velocity}\\(mm)$\downarrow$} \\
\midrule
Clean-only AE
& \textbf{2.339} & \textbf{3.887} & \textbf{0.407}
& \textbf{1.872} & \textbf{4.500} & \textbf{0.216} \\
Random-mask AE
& \underline{2.440} & \underline{5.392} & \underline{0.423}
& \underline{3.339} & \underline{7.469} & 0.924 \\
\rowcolor{orange!12}
\textbf{Structured-mask AE (Ours)}
& 3.392 & 6.199 & 0.455
& 3.807 & 8.899 & \underline{0.740} \\
\bottomrule
\end{tabular}%
}
\end{table*}

\begin{table*}[t]
\centering
\caption{
\textbf{Clean-motion reconstruction after Stage~2 on RoMo.}
We evaluate the refined tokenizers used for downstream
generation on unmasked inputs.
Metrics and notation follow Table~\ref{tab:ae_clean_stage1}.
}
\label{tab:ae_clean_stage2}
\scriptsize
\setlength{\tabcolsep}{5pt}
\renewcommand{\arraystretch}{1.1}
\resizebox{0.8\textwidth}{!}{%
\begin{tabular}{lcccccc}
\toprule
\textbf{Tokenizer}
& \makecell{\textbf{MPJPE}\\local (mm)$\downarrow$}
& \makecell{\textbf{MPJPE}\\world (mm)$\downarrow$}
& \makecell{\textbf{Rotation}\\(deg)$\downarrow$}
& \makecell{\textbf{Root path}\\(mm)$\downarrow$}
& \makecell{\textbf{Final pos.}\\(mm)$\downarrow$}
& \makecell{\textbf{Velocity}\\(mm)$\downarrow$} \\
\midrule
Clean-only AE
& \textbf{1.876} & \textbf{2.400} & \underline{0.388}
& \textbf{0.800} & \textbf{1.706} & \textbf{0.207} \\
Random-mask AE
& \underline{1.897} & 3.392 & \textbf{0.360}
& 1.811 & 4.134 & 0.833 \\
\rowcolor{orange!12}
\textbf{Structured-mask AE (Ours)}
& 2.148 & \underline{3.127} & 0.394
& \underline{1.284} & \underline{2.822} & \underline{0.642} \\
\bottomrule
\end{tabular}%
}
\end{table*}

\begin{table*}[t]
\centering
\caption{
\textbf{Masked-motion reconstruction after Stage~2.}
All tokenizers are evaluated on the same four masking
families using their refined decoders, without the
latent generator.
We report MPJPE (mm) over hidden regions only.
Lower is better; \textbf{bold} and \underline{underlined}
values denote the best and second-best results, respectively.
}
\label{tab:ae_masked_stage2}
\scriptsize
\setlength{\tabcolsep}{5pt}
\renewcommand{\arraystretch}{1.1}
\resizebox{0.8\textwidth}{!}{%
\begin{tabular}{lcccc}
\toprule
\textbf{Tokenizer}
& \textbf{Anatomical tube}$\downarrow$
& \textbf{Temporal stride}$\downarrow$
& \textbf{Feature-group}$\downarrow$
& \textbf{Mixed}$\downarrow$ \\
\midrule
Clean-only AE
& 257.8 & 207.4 & \underline{273.1} & 249.4 \\
Random-mask AE
& \underline{201.7} & \underline{9.3}
& 306.7 & \underline{199.8} \\
\rowcolor{orange!12}
\textbf{Structured-mask AE (Ours)}
& \textbf{46.2} & \textbf{7.0}
& \textbf{158.7} & \textbf{43.8} \\
\bottomrule
\end{tabular}%
}
\end{table*}

\paragraph{Reconstruction fidelity and condition encoding.}
Tables~\ref{tab:ae_clean_stage1} and~\ref{tab:ae_clean_stage2}
evaluate reconstruction from complete inputs before and after
decoder refinement, independently of the latent generator.
The clean-only tokenizer achieves the lowest errors on all
Stage~1 metrics and five of six Stage~2 metrics, consistent with
its objective being dedicated to complete-motion reconstruction.
The random-mask tokenizer remains competitive in local joint
reconstruction and achieves the lowest Stage~2 rotation error.
One plausible explanation is that dependencies between connected
joints and neighboring frames allow independently masked entries
to be recovered from nearby observations. Structured masks remove
coordinated sets of observations and may therefore require broader
context. The tables are consistent with this interpretation but
do not directly establish which dependencies each tokenizer uses.

The structured-mask tokenizer has higher clean-reconstruction errors
than the clean-only tokenizer, yet achieves stronger downstream
generation in Table~\ref{tab:ablation_masking_main}.
Thus, clean reconstruction alone is insufficient for tokenizer
selection. This observation complements VA-VAE~\citep{yao2025reconstruction},
which identifies an optimization difficulty when increasing latent
dimensionality improves reconstruction but makes diffusion modeling
more demanding. Its latent alignment strategy improves both
objectives, emphasizing the importance of latent structure rather
than an inherent conflict between reconstruction and generation.
Our comparison instead varies masking at a fixed tokenizer
architecture, providing complementary evidence in the motion domain.

Table~\ref{tab:ae_masked_stage2} evaluates all refined tokenizers
on the same four masking families, measuring MPJPE only over hidden
regions. Structured masking achieves the lowest errors throughout,
reducing anatomical-tube and mixed-mask errors from 201.7 to 46.2\,mm
and from 199.8 to 43.8\,mm, respectively, relative to random masking.
Random masking performs well under temporal stride masks but transfers
poorly to the other structured gaps. Effective recovery under one
masking pattern therefore does not ensure reliable use of
heterogeneous partial observations.

Together, these results support our tokenizer's dual role in encoding
complete generation targets and partial-motion conditions with a
shared encoder. Hidden-region reconstruction assesses how the encoder
and decoder jointly use partial observations; it does not isolate
encoder quality. Combined with the downstream ablations, the evidence
favors balancing clean reconstruction with effective condition
encoding rather than selecting the tokenizer by unmasked error alone.

\paragraph{Decoder refinement.}
Table~\ref{tab:supp_reconstruction} isolates decoder refinement
for the default tokenizer on 2,048 test clips per dataset, using
deterministic center crops of at most 128 frames.
With the encoder fixed, Stage~2 improves local position, rotation,
and root-path reconstruction on both datasets.
For example, local MPJPE decreases from 3.392 to 2.148\,mm on RoMo
and from 3.160 to 2.063\,mm on MotionMillion.
These improvements show that clean reconstruction can be refined
without changing the encoder's representation. They measure
reconstruction from complete inputs and do not establish an
independent improvement in masked completion or generated motion.

\begin{table*}[tbp]
\centering
\caption{\textbf{Effect of decoder refinement.}
MPJPE uses all valid joints and frames of clean inputs, and path error
uses all valid frames. All metrics are lower-is-better; bold marks the
better stage within each dataset.}
\label{tab:supp_reconstruction}
\scriptsize
\setlength{\tabcolsep}{3pt}
\renewcommand{\arraystretch}{1.1}
\resizebox{0.8\textwidth}{!}{%
\begin{tabular}{lcccccc}
\toprule
& \multicolumn{3}{c}{\textbf{RoMo}}
& \multicolumn{3}{c}{\textbf{MotionMillion}} \\
\cmidrule(lr){2-4}
\cmidrule(lr){5-7}
\textbf{Stage}
& \makecell{\textbf{Local MPJPE}\\(mm)$\downarrow$}
& \makecell{\textbf{Rotation error}\\($^\circ$)$\downarrow$}
& \makecell{\textbf{Path error}\\(mm)$\downarrow$}
& \makecell{\textbf{Local MPJPE}\\(mm)$\downarrow$}
& \makecell{\textbf{Rotation error}\\($^\circ$)$\downarrow$}
& \makecell{\textbf{Path error}\\(mm)$\downarrow$} \\
\midrule
Stage 1 & 3.392 & 0.455 & 3.807 & 3.160 & 0.460 & 4.358 \\
\rowcolor{orange!12}
Stage 2 & \textbf{2.148} & \textbf{0.394} & \textbf{1.284}
& \textbf{2.063} & \textbf{0.363} & \textbf{1.909} \\
\bottomrule
\end{tabular}%
}
\end{table*}

\paragraph{Preservation of supplied observations.}
Table~\ref{tab:supp_observed} complements hidden-region and
distributional metrics with errors on supplied observations.
The full model achieves the lowest point estimates on all seven
RoMo tasks, using the same frozen tokenizer as both ablated variants.
The differences are clearest for pose-conditioned generation:
removing the observation map increases P2M and TP2M errors from
5.06 and 4.63\,mm to 6.25 and 5.92\,mm, respectively.
Removing the observation loss also increases both errors.
This supports using explicit observation information and supervision
to preserve conditions during generation. Some differences, such as
those for extrapolation, are small and should not be interpreted as
statistically established improvements. All errors are measured on
decoded outputs without copying the supplied observations back.

\begin{table*}[tbp]
\centering
\caption{\textbf{Observation preservation on RoMo.}
The first six columns report observed-joint MPJPE; Trajectory reports
observed root-$xz$ error. All values are in millimeters on 512 clips
per task. All variants use the same frozen tokenizer. \textbf{Bold} and
\underline{underlined} values mark the lowest and second-lowest point estimates,
not statistical significance.}
\label{tab:supp_observed}
\scriptsize
\setlength{\tabcolsep}{3pt}
\renewcommand{\arraystretch}{1.1}
\resizebox{0.8\textwidth}{!}{%
\begin{tabular}{lccccccc}
\toprule
\textbf{Variant}
& \textbf{FLF}$\downarrow$ & \textbf{FI}$\downarrow$
& \textbf{Extrap.}$\downarrow$ & \textbf{Partial}$\downarrow$
& \textbf{P2M}$\downarrow$ & \textbf{TP2M}$\downarrow$
& \textbf{Trajectory}$\downarrow$ \\
\midrule
w/o observation map
& \underline{5.50} & 4.32  & \underline{2.92} & \underline{3.56}
& \underline{6.25} & \underline{5.92} & 8.06 \\
w/o $\mathcal{L}_{\mathrm{obs}}$
& 5.78 & \underline{3.78} & 2.98 & 3.74
& 6.40 & 5.94 & \underline{7.93} \\
\rowcolor{orange!12}
\textbf{MotionMaestro (Ours)}
& \textbf{4.85} & \textbf{3.70} & \textbf{2.91} & \textbf{3.49}
& \textbf{5.06} & \textbf{4.63} & \textbf{7.51} \\
\bottomrule
\end{tabular}%
}
\end{table*}

\paragraph{Sampling variability.}
Table~\ref{tab:supp_variability} repeats RoMo T2M sampling with
seeds 0, 1, and 2.
The full model has the lowest mean FID among the compared variants;
the paired differences are computed against its result at each seed.
The clean-only and random-mask tokenizers have larger mean FID gaps
than the two observation-related ablations.
The latter have small positive mean differences, so these results
do not support a T2M advantage from removing the observation map or
observation loss. Their primary role is more directly assessed by
condition-preservation metrics in Table~\ref{tab:supp_observed}.
This analysis measures sampling variation for fixed checkpoints,
not variation across independent training runs; three seeds are
insufficient for broad significance claims.
The main comparisons retain their seed-0 results, and other results
use a single sampling seed unless stated otherwise.

\begin{table*}[htbp]
\centering
\caption{\textbf{RoMo T2M sampling variability.}
Mean and sample standard deviation over three sampling seeds.
Paired $\Delta$ subtracts the full-model FID at the same seed;
positive values indicate higher FID.}
\label{tab:supp_variability}
\scriptsize
\setlength{\tabcolsep}{5pt}
\renewcommand{\arraystretch}{1.1}
\resizebox{0.55\textwidth}{!}{%
\begin{tabular}{lcc}
\toprule
\textbf{Variant} & \textbf{FID}$\downarrow$
& \textbf{Paired $\Delta$ vs. full model} \\
\midrule
Clean sequences only & 17.910$\pm$0.360 & +3.335$\pm$0.161 \\
Random masking & 15.928$\pm$0.576 & +1.352$\pm$0.749 \\
w/o observation map & 14.918$\pm$0.366 & +0.342$\pm$0.496 \\
w/o $\mathcal{L}_{\mathrm{obs}}$ & 15.056$\pm$0.047 & +0.480$\pm$0.324 \\
\rowcolor{orange!12}
\textbf{MotionMaestro (Ours)} & 14.575$\pm$0.283 & -- \\
\bottomrule
\end{tabular}%
}
\end{table*}

\section{Additional Experiments}
\label{sec:supp_additional_results}

We first compare both MotionMaestro variants across the ten tasks
and present additional qualitative results for generation and editing.
We then examine sensitivity to the temporal and spatial extent of
observations. Physical diagnostics and text-guidance sweeps provide
complementary assessments of motion statistics and text alignment.

\paragraph{Comparison with MotionMaestro-5B.}
Tables~\ref{tab:main_romo_5b} and~\ref{tab:main_mm_5b} extend the
main comparisons to MotionMaestro-5B.
Relative to the default model, the 5B variant improves the reported
metric on nine of ten tasks in each dataset; the exceptions are
T2M on RoMo and unconditional generation on MotionMillion.
The gains are particularly visible in partial completion and
long-sequence generation. On MotionMillion, the 5B variant also
improves P2M and TP2M, while MotionLab retains the lowest
trajectory-conditioned FID.
The reported inference time changes little between our variants,
whereas memory increases from 11.7 to 30.6\,GB.
These results characterize the larger model configuration, rather
than isolating parameter count: in addition to greater width,
MotionMaestro-5B uses pooled CLIP text conditioning
(Section~\ref{app:mm_implementation}).

\begin{figure}[bt]
    \centering
    \includegraphics[width=\textwidth]{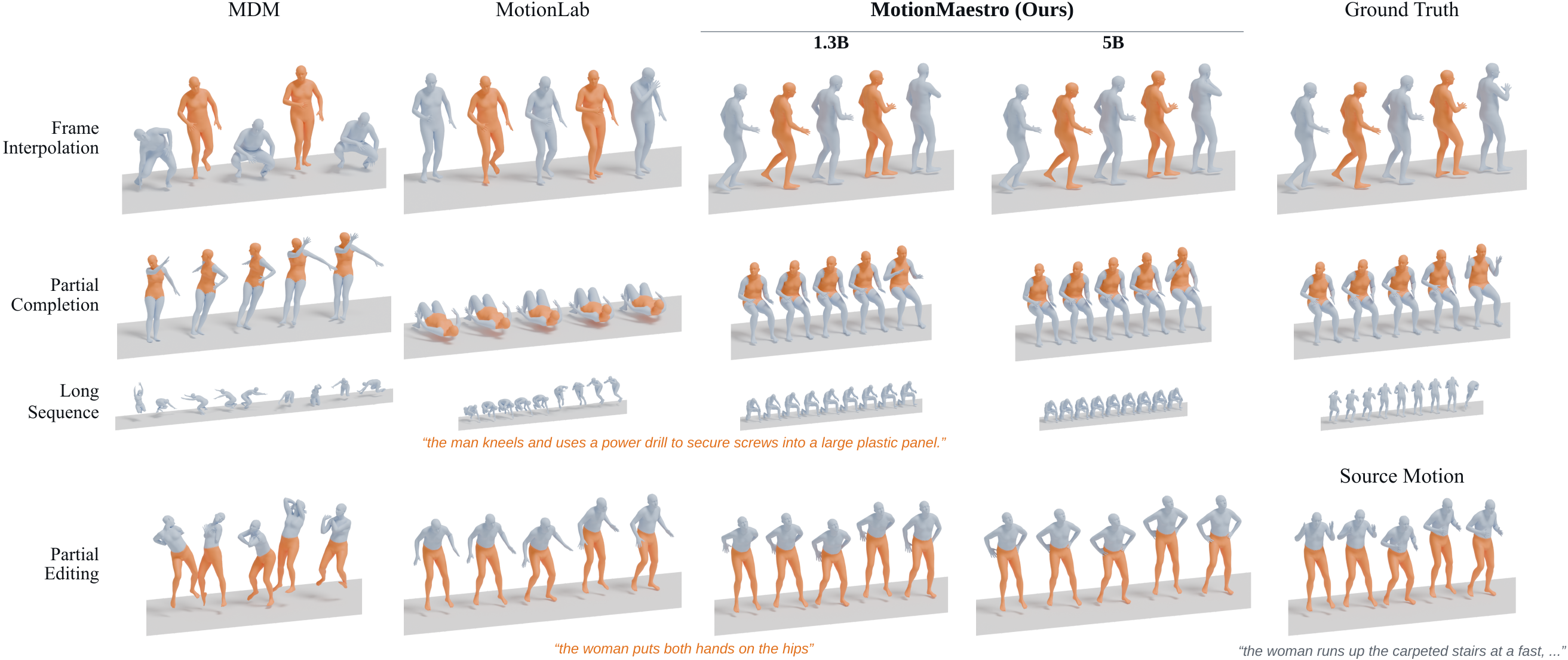}
    \caption{\textbf{Additional qualitative comparison (1/5).}
    We compare MDM, MotionLab, and the 1.3B and 5B variants of
    MotionMaestro. Rows show frame interpolation, partial completion,
    long-sequence generation, and text-guided partial editing,
    respectively. \textcolor{motionorange}{Orange} denotes supplied motion observations.
    The rightmost column shows ground-truth motions for the first
    three rows and the source motion for partial editing.
    The editing prompt requests placing both hands on the hips.}
    \label{fig:supp_qual_1}
\end{figure}

\begin{figure}[p]
    \centering
    \includegraphics[width=\textwidth]{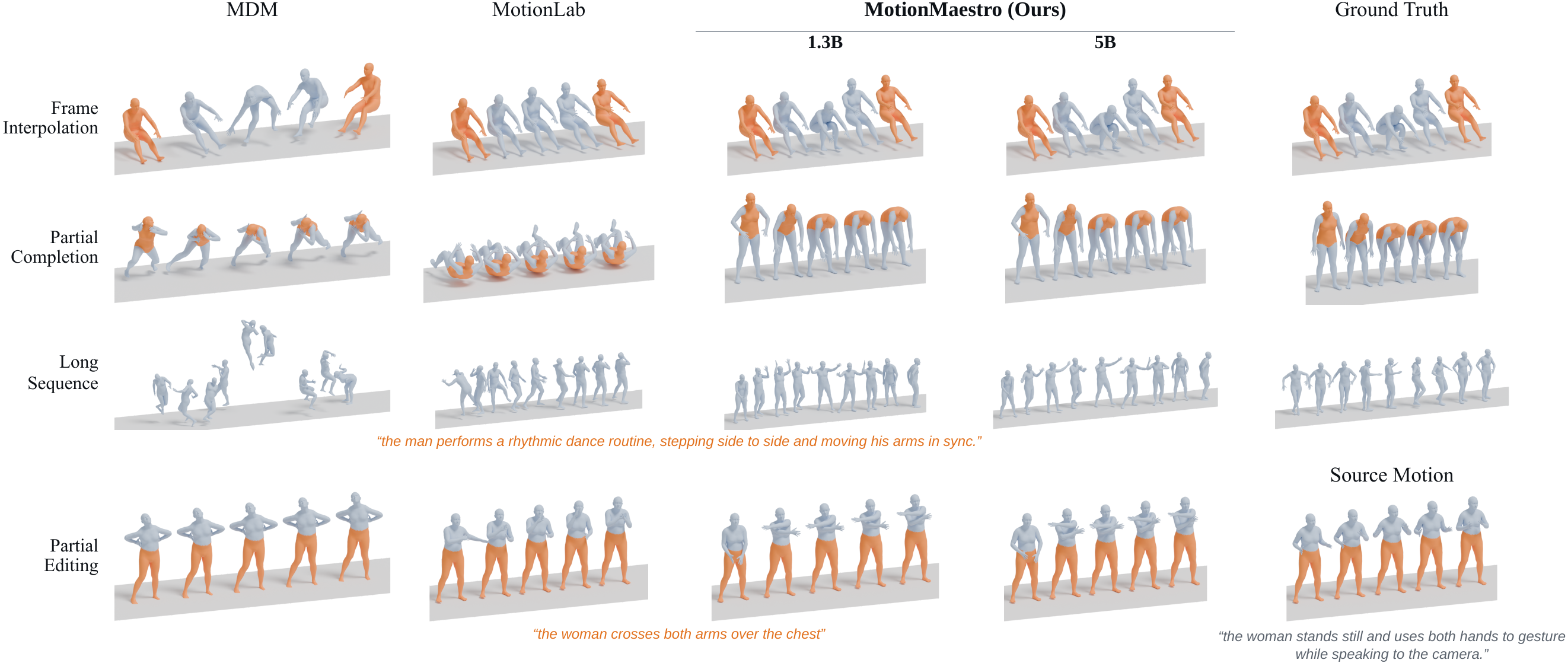}
    \caption{\textbf{Additional qualitative comparison (2/5).}
    Methods, row order, and color conventions follow
    Figure~\ref{fig:supp_qual_1}.
    The editing prompt requests crossing both arms over the chest
    while retaining the supplied lower-body motion.}
    \label{fig:supp_qual_2}
\end{figure}

\begin{figure}[p]
    \centering
    \includegraphics[width=\textwidth]{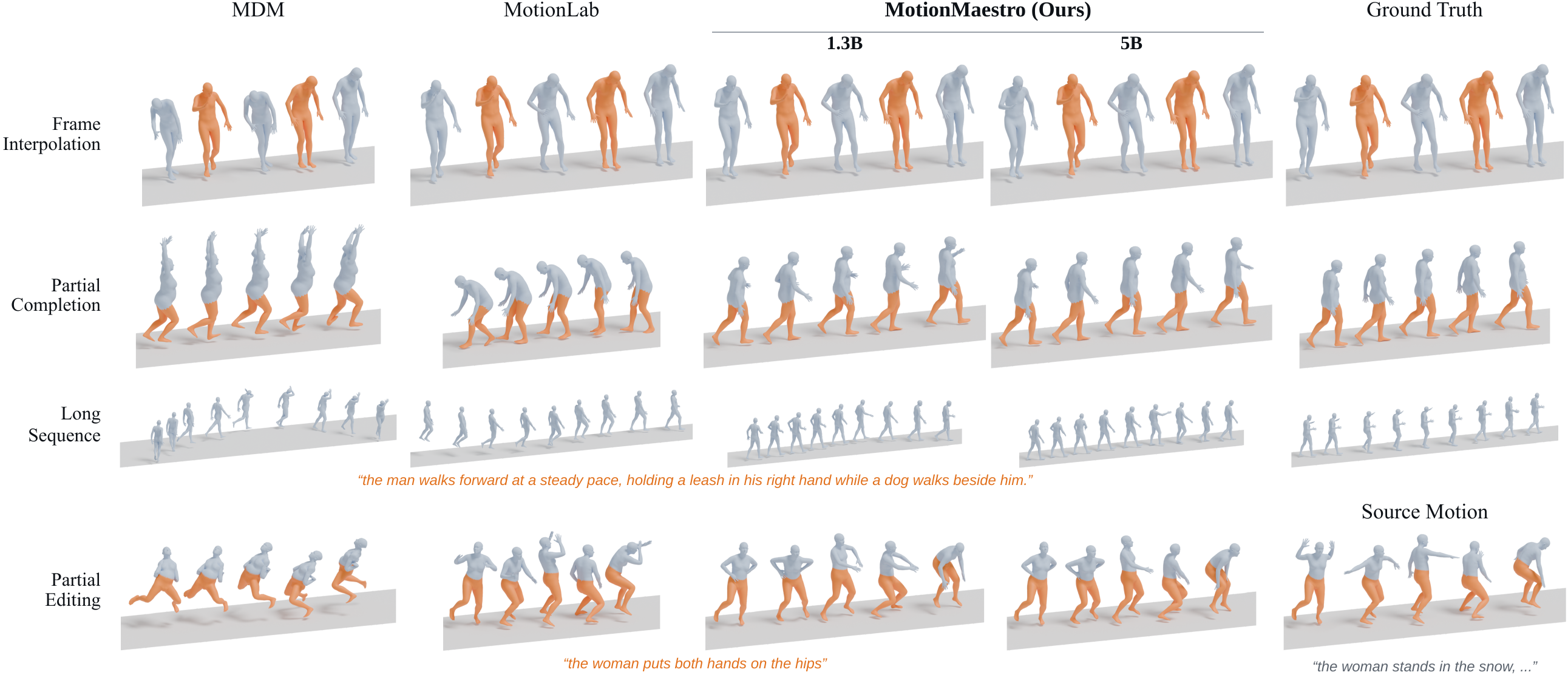}
    \caption{\textbf{Additional qualitative comparison (3/5).}
    Methods, row order, and color conventions follow
    Figure~\ref{fig:supp_qual_1}.
    The editing prompt requests placing both hands on the hips
    while retaining the supplied lower-body motion.}
    \label{fig:supp_qual_3}
\end{figure}

\begin{figure}[p]
    \centering
    \includegraphics[width=\textwidth]{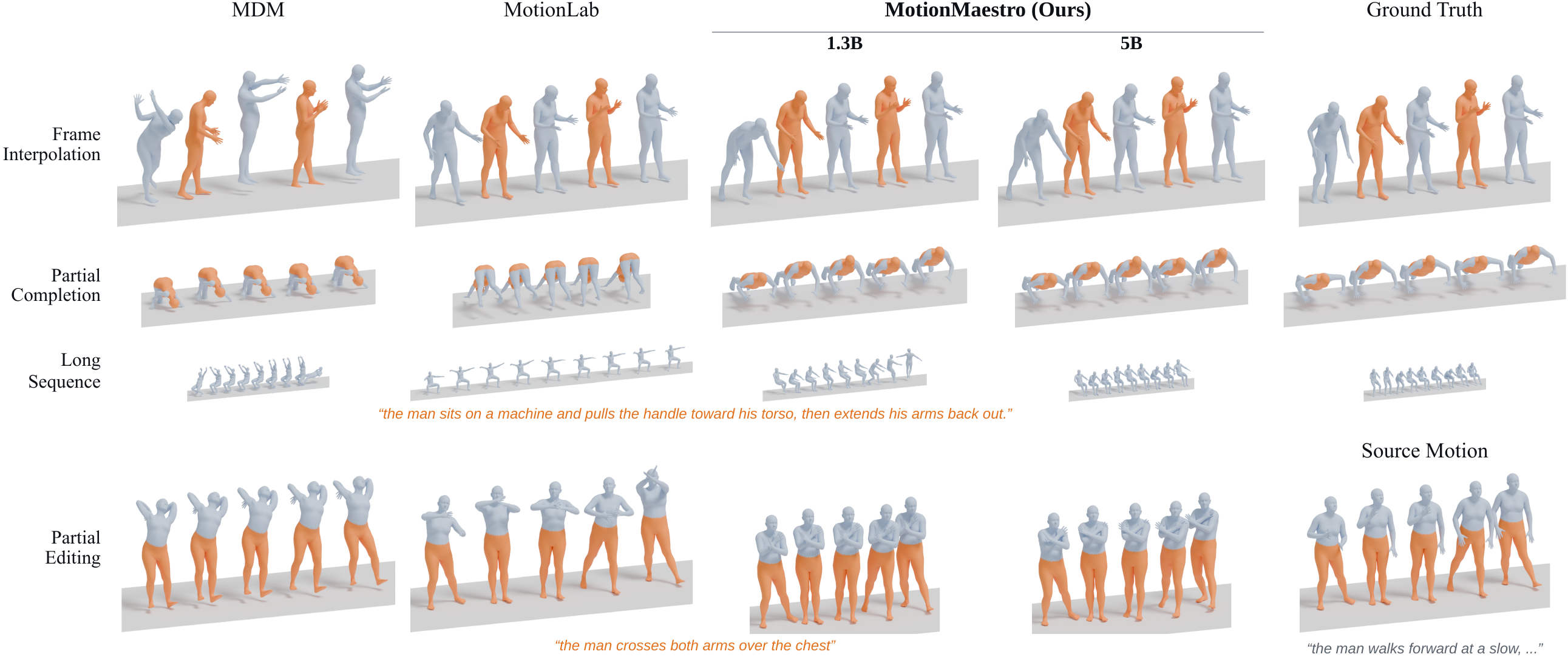}
    \caption{\textbf{Additional qualitative comparison (4/5).}
    Methods, row order, and color conventions follow
    Figure~\ref{fig:supp_qual_1}.
    The editing prompt requests crossing both arms over the chest
    while retaining the supplied lower-body motion.}
    \label{fig:supp_qual_4}
\end{figure}

\begin{figure}[p]
    \centering
    \includegraphics[width=\textwidth]{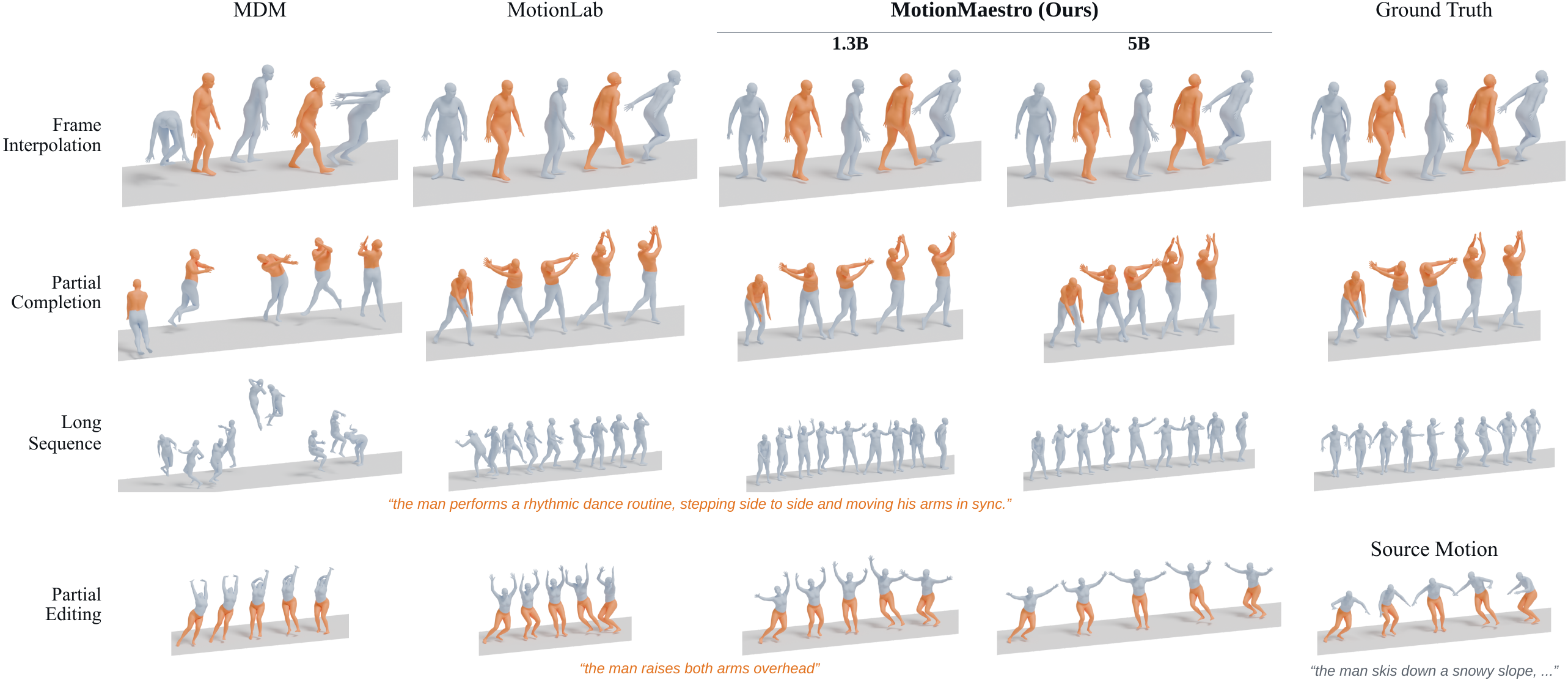}
    \caption{\textbf{Additional qualitative comparison (5/5).}
    Methods, row order, and color conventions follow
    Figure~\ref{fig:supp_qual_1}.
    The editing prompt requests raising both arms overhead
    while retaining the supplied lower-body motion.}
    \label{fig:supp_qual_5}
\end{figure}

\paragraph{Qualitative comparison.}
Figures~\ref{fig:supp_qual_1}--\ref{fig:supp_qual_5} compare
MDM, MotionLab, and both MotionMaestro variants on frame interpolation,
partial completion, long-sequence generation, and text-guided partial editing.
In the interpolation examples, both variants recover intermediate poses
that closely resemble the reference while respecting the observed keyframes.
For partial completion, they generate missing body movements compatible
with the observed regions, avoiding the pronounced posture deviations
visible in several baseline outputs.
The long-sequence examples further illustrate coherent pose progression
for actions such as walking, dancing, and seated pulling.
For partial editing, MotionMaestro combines source lower-body observations
with new instructions, such as placing the hands on the hips or crossing
the arms over the chest, without editing-specific training or fine-tuning.
Instruction adherence varies across examples, and the edited movements
do not always fully realize the requested pose.
These visualizations complement the quantitative evaluation by illustrating
both the capabilities and remaining limitations of the shared conditioning
interface.

\begin{table*}[t]
\centering
\caption{
\textbf{Unified motion generation including MotionMaestro-5B on RoMo.}
FI and Partial report MPJPE (mm) over unobserved frames and joints,
respectively; the remaining tasks report FID.
Lower is better; \textbf{bold} and \underline{underlined} mark the best
and second-best results per task.}
\label{tab:main_romo_5b}
\scriptsize
\setlength{\tabcolsep}{3pt}
\renewcommand{\arraystretch}{1.1}
\resizebox{\textwidth}{!}{%
\begin{tabular}{lccccccccccc}
\toprule
\textbf{Method} & \textbf{Venue}
& \makecell{\textbf{Uncond.}\\FID$\downarrow$} & \makecell{\textbf{T2M}\\FID$\downarrow$} & \makecell{\textbf{P2M}\\FID$\downarrow$} & \makecell{\textbf{TP2M}\\FID$\downarrow$} & \makecell{\textbf{FLF}\\FID$\downarrow$} & \makecell{\textbf{FI}\\MPJPE$\downarrow$} & \makecell{\textbf{Extrap.}\\FID$\downarrow$} & \makecell{\textbf{Partial}\\MPJPE$\downarrow$} & \makecell{\textbf{Trajectory}\\FID$\downarrow$} & \makecell{\textbf{Long}\\FID$\downarrow$} \\
\midrule
MDM~\citep{tevet2022mdm} & ICLR2023
& 507.6 & 450.7 & 518.4 & 456.1 & 523.3 & 117.6 & 342.8 & 278.9 & -- & 510.9 \\
OmniControl~\citep{xie2024omnicontrol} & ICLR2024
& 429.1 & 199.4 & 771.0 & 312.8 & 753.2 & 875.9 & 740.3 & 627.3 & 283.7 & 398.6 \\
MotionLab~\citep{guo2025motionlab} & ICCV2025
& 76.7 & 101.5 & 53.5 & 46.5 & 30.0 & 12.7 & 28.1 & 71.8 & 89.3 & 123.5 \\
\rowcolor{orange!12}
\textbf{MotionMaestro (Ours)} & --
& \underline{20.4} & \textbf{14.9} & \underline{34.8} & \underline{28.7} & \underline{20.8} & \underline{6.7} & \underline{19.4} & \underline{61.0} & \underline{59.2} & \underline{26.2} \\
\rowcolor{orange!12}
\textbf{MotionMaestro-5B (Ours)} & --
& \textbf{20.3} & \underline{17.2} & \textbf{31.3} & \textbf{28.6} & \textbf{18.4} & \textbf{5.8} & \textbf{17.9} & \textbf{47.2} & \textbf{59.0} & \textbf{19.0} \\
\bottomrule
\end{tabular}%
}
\end{table*}

\begin{table*}[t]
\centering
\caption{
\textbf{Unified motion generation and computational cost
on MotionMillion.}
Evaluation settings, metrics, and notation follow
Table~\ref{tab:main_romo_5b}.
We additionally report computational cost.
}
\label{tab:main_mm_5b}
\scriptsize
\setlength{\tabcolsep}{3pt}
\renewcommand{\arraystretch}{1.1}
\resizebox{\textwidth}{!}{%
\begin{tabular}{lccccccccccccc}
\toprule
\textbf{Method}
& \makecell{\textbf{Uncond.}\\FID$\downarrow$} & \makecell{\textbf{T2M}\\FID$\downarrow$} & \makecell{\textbf{P2M}\\FID$\downarrow$} & \makecell{\textbf{TP2M}\\FID$\downarrow$} & \makecell{\textbf{FLF}\\FID$\downarrow$} & \makecell{\textbf{FI}\\MPJPE$\downarrow$} & \makecell{\textbf{Extrap.}\\FID$\downarrow$} & \makecell{\textbf{Partial}\\MPJPE$\downarrow$} & \makecell{\textbf{Trajectory}\\FID$\downarrow$} & \makecell{\textbf{Long}\\FID$\downarrow$}
& \makecell{\textbf{Inference}\\Time (s)}
& \makecell{\textbf{Memory}\\GB}
& \makecell{\textbf{Params}\\Billion (B)} \\
\midrule
MDM & 427.7 & 297.6 & 523.8 & 334.4 & 530.6 & 122.8 & 370.4 & 206.8 & -- & 336.2 & 4.69 & 0.4 & 0.081 \\
OmniControl & 376.3 & 153.9 & 366.8 & 193.8 & 365.5 & 375.4 & 551.3 & 349.6 & 276.6 & 177.9 & 61.58 & 0.5 & 0.102 \\
MotionLab & 135.1 & 61.4 & 74.4 & \underline{49.4} & 50.7 & 12.3 & 38.8 & 71.1 & \textbf{94.4} & 108.6 & 0.97 & 2.6 & 0.367 \\
\rowcolor{orange!12}
\textbf{MotionMaestro} & \textbf{67.2} & \underline{36.2} & \underline{74.3} & 52.8 & \underline{44.5} & \underline{3.8} & \underline{36.9} & \underline{61.3} & 118.2 & \underline{46.4} & 1.28 & 11.7 & 1.285 \\
\rowcolor{orange!12}
\textbf{MotionMaestro-5B} & \underline{67.9} & \textbf{34.7} & \textbf{51.3} & \textbf{42.8} & \textbf{32.5} & \textbf{3.2} & \textbf{24.3} & \textbf{45.2} & \underline{109.5} & \textbf{35.8} & 1.31 & 30.6 & 5.096 \\
\bottomrule
\end{tabular}
}
\end{table*}

\paragraph{Frame interpolation with varying observation stride.}
Table~\ref{tab:supp_fi_stride} extends the main-paper $s=4$ setting
to denser and sparser observations on both datasets.
For both MotionMaestro variants, hidden-region MPJPE increases as
the stride grows, reflecting the greater uncertainty between anchors.
Nevertheless, the default model outperforms all three baselines in
both MPJPE and FID at every reported stride on each dataset.
MotionMaestro-5B further reduces MPJPE throughout and improves FID
in all cases except RoMo at $s=2$.

\begin{table*}[t]
\centering
\caption{
\textbf{Frame interpolation with varying observation stride on RoMo and MotionMillion.}
MPJPE (mm) is computed over unobserved frames; FID uses complete outputs.
Lower is better; \textbf{bold} and \underline{underlined}
mark the best and second-best results per column.
}
\label{tab:supp_fi_stride}
\scriptsize
\setlength{\tabcolsep}{3pt}
\renewcommand{\arraystretch}{1.1}
\resizebox{\textwidth}{!}{%
\begin{tabular}{lcccccccccccccccc}
\toprule
\multirow{3}{*}{\textbf{Method}}
& \multicolumn{8}{c}{\textbf{RoMo}}
& \multicolumn{8}{c}{\textbf{MotionMillion}} \\
\cmidrule(lr){2-9}
\cmidrule(lr){10-17}
& \multicolumn{2}{c}{$s{=}2$}
& \multicolumn{2}{c}{$s{=}8$}
& \multicolumn{2}{c}{$s{=}16$}
& \multicolumn{2}{c}{$s{=}32$}
& \multicolumn{2}{c}{$s{=}2$}
& \multicolumn{2}{c}{$s{=}8$}
& \multicolumn{2}{c}{$s{=}16$}
& \multicolumn{2}{c}{$s{=}32$} \\
\cmidrule(lr){2-3}
\cmidrule(lr){4-5}
\cmidrule(lr){6-7}
\cmidrule(lr){8-9}
\cmidrule(lr){10-11}
\cmidrule(lr){12-13}
\cmidrule(lr){14-15}
\cmidrule(lr){16-17}
& MPJPE$\downarrow$ & FID$\downarrow$
& MPJPE$\downarrow$ & FID$\downarrow$
& MPJPE$\downarrow$ & FID$\downarrow$
& MPJPE$\downarrow$ & FID$\downarrow$
& MPJPE$\downarrow$ & FID$\downarrow$
& MPJPE$\downarrow$ & FID$\downarrow$
& MPJPE$\downarrow$ & FID$\downarrow$
& MPJPE$\downarrow$ & FID$\downarrow$ \\
\midrule
MDM
& 59.3 & 61.9 & 252.2 & 282.5
& 375.2 & 403.6 & 431.3 & 464.0
& 75.4 & 128.7 & 203.5 & 301.0
& 285.9 & 401.1 & 330.4 & 454.8 \\
OmniControl
& 803.2 & 747.8 & 929.2 & 782.7
& 943.3 & 776.1 & 930.6 & 745.8
& 383.9 & 503.6 & 372.0 & 473.5
& 373.9 & 457.7 & 373.6 & 426.0 \\
MotionLab
& 11.1 & 10.7 & 18.9 & 14.1
& 35.4 & 21.4 & 57.5 & 30.4
& 11.6 & 9.4 & 15.7 & 11.4
& 26.1 & 16.9 & 47.0 & 27.8 \\
\rowcolor{orange!12}
\textbf{MotionMaestro}
& \underline{4.8} & \textbf{3.9}
& \underline{12.7} & \underline{4.7}
& \underline{30.1} & \underline{12.9}
& \underline{54.1} & \underline{18.6}
& \underline{2.8} & \underline{1.5}
& \underline{7.5} & \underline{1.8}
& \underline{20.3} & \underline{5.9}
& \underline{43.1} & \underline{15.2} \\
\rowcolor{orange!12}
\textbf{MotionMaestro-5B}
& \textbf{4.3} & \underline{4.3}
& \textbf{11.1} & \textbf{4.4}
& \textbf{26.0} & \textbf{10.1}
& \textbf{47.8} & \textbf{15.6}
& \textbf{2.3} & \textbf{1.0}
& \textbf{6.3} & \textbf{1.6}
& \textbf{15.9} & \textbf{4.5}
& \textbf{36.0} & \textbf{11.4} \\
\bottomrule
\end{tabular}%
}
\end{table*}

\paragraph{Partial completion under different observed body parts.}
Table~\ref{tab:supp_partial} evaluates spatially restricted observations,
with the observed body parts specified separately for each dataset.
MotionMaestro-5B achieves the lowest hidden-joint MPJPE under all six
settings and the lowest FID in four; MotionLab retains the lowest
FID for one-arm conditioning on both datasets.
The default model improves both metrics over prior methods for
torso and head$+$hands on RoMo, and for upper
on MotionMillion. However, MotionLab performs better than the
default model under the two sparser MotionMillion presets.
The gains therefore depend on the available observations, and
lower reconstruction error does not always coincide with lower FID.
OmniControl consumes only supported joints from its six-joint control
set, so different nominal masks may yield identical hints for it.

\begin{table*}[t]
\centering
\caption{
\textbf{Partial completion under different observed body parts
on RoMo and MotionMillion.}
We report MPJPE (mm) over unobserved joints and FID.
Observed body parts are specified separately for each dataset.
Lower is better; \textbf{bold} and \underline{underlined}
mark the best and second-best results, respectively.
}
\label{tab:supp_partial}
\scriptsize
\setlength{\tabcolsep}{3pt}
\renewcommand{\arraystretch}{1.1}
\resizebox{\textwidth}{!}{%
\begin{tabular}{lcccccccccccc}
\toprule
\multirow{3}{*}{\textbf{Method}}
& \multicolumn{6}{c}{\textbf{RoMo}}
& \multicolumn{6}{c}{\textbf{MotionMillion}} \\
\cmidrule(lr){2-7}
\cmidrule(lr){8-13}
& \multicolumn{2}{c}{torso}
& \multicolumn{2}{c}{head$+$hands}
& \multicolumn{2}{c}{one arm}
& \multicolumn{2}{c}{upper}
& \multicolumn{2}{c}{head$+$hands}
& \multicolumn{2}{c}{one arm} \\
\cmidrule(lr){2-3}
\cmidrule(lr){4-5}
\cmidrule(lr){6-7}
\cmidrule(lr){8-9}
\cmidrule(lr){10-11}
\cmidrule(lr){12-13}
& MPJPE$\downarrow$ & FID$\downarrow$
& MPJPE$\downarrow$ & FID$\downarrow$
& MPJPE$\downarrow$ & FID$\downarrow$
& MPJPE$\downarrow$ & FID$\downarrow$
& MPJPE$\downarrow$ & FID$\downarrow$
& MPJPE$\downarrow$ & FID$\downarrow$ \\
\midrule
MDM
& 301.4 & 553.8 & 346.0 & 516.8 & 313.5 & 561.4
& 206.8 & 401.9 & 247.9 & 469.3 & 230.6 & 470.0 \\
OmniControl
& 535.3 & 477.5 & 627.3 & 631.5 & 297.4 & 293.1
& 349.6 & 448.5 & 349.6 & 448.5 & 252.5 & 287.0 \\
MotionLab
& 378.1 & 379.3 & 73.8 & 35.3 & 81.2 & \textbf{32.7}
& 71.1 & 24.3
& \underline{68.6} & \underline{47.7}
& \underline{74.7} & \textbf{40.5} \\
\rowcolor{orange!12}
\textbf{MotionMaestro}
& \underline{109.3} & \underline{44.7}
& \underline{68.5} & \underline{31.8}
& \underline{79.5} & 39.0
& \underline{61.3} & \underline{23.5}
& 73.7 & 67.0
& 79.7 & 62.2 \\
\rowcolor{orange!12}
\textbf{MotionMaestro-5B}
& \textbf{95.8} & \textbf{42.2}
& \textbf{56.7} & \textbf{30.3}
& \textbf{68.2} & \underline{33.6}
& \textbf{45.2} & \textbf{13.8}
& \textbf{60.6} & \textbf{46.4}
& \textbf{64.6} & \underline{46.3} \\
\bottomrule
\end{tabular}%
}
\end{table*}

\paragraph{Physical plausibility.}
Table~\ref{tab:physical} evaluates the default model directly from
joint positions, complementing the learned feature-space metrics.
MotionMaestro is closest to the real-motion reference on all six
reported measures in both datasets, among the compared methods.
The agreement covers contact-related statistics and motion
derivatives, supporting the distributional results with geometric
diagnostics. Residual differences remain: for example, the foot-skate
ratio is 0.841 versus 0.611 for real RoMo motions.
Closeness to reference statistics is the criterion used in this
table; it does not imply that the outputs are free of physical
artifacts or that real-motion statistics are ideal physical targets.

\begin{table*}[t]
\centering
\caption{
\textbf{Physical plausibility on RoMo and MotionMillion.}
Metrics are computed directly from joint positions without a learned evaluator.
Foot skate measures the fraction of contact frames with foot movement.
Penetration and float measure the percentage of frames whose lowest joint
is more than 10\,mm below or 15\,mm above the floor, respectively.
\textbf{Bold} and \underline{underlined} mark the closest and second-closest
generated results to ground truth per metric and dataset.
Mean acceleration is displayed with a $10^{-3}$ scale factor.
}
\label{tab:physical}
\scriptsize
\setlength{\tabcolsep}{3pt}
\renewcommand{\arraystretch}{1.1}
\resizebox{\textwidth}{!}{%
\begin{tabular}{lcccccccccccc}
\toprule
& \multicolumn{6}{c}{\textbf{RoMo}}
& \multicolumn{6}{c}{\textbf{MotionMillion}} \\
\cmidrule(lr){2-7}
\cmidrule(lr){8-13}
\textbf{Method}
& \makecell{Foot skate\\ratio}
& \makecell{GMD skate\\(mm/s)}
& \makecell{Penetration\\(\%)}
& \makecell{Float\\(\%)}
& \makecell{Peak\\jerk}
& \makecell{Accel.\\mean}
& \makecell{Foot skate\\ratio}
& \makecell{GMD skate\\(mm/s)}
& \makecell{Penetration\\(\%)}
& \makecell{Float\\(\%)}
& \makecell{Peak\\jerk}
& \makecell{Accel.\\mean} \\
\midrule
\rowcolor{black!5}
Ground Truth
& 0.611 & 76.5 & 0.77 & 25.31 & 561.5 & 4.83
& 0.509 & 63.1 & 0.66 & 25.42 & 225.9 & 2.17 \\
MDM~\citep{tevet2022mdm}
& 0.961 & 163.5 & 1.59 & 40.17 & 2306.8 & 17.76
& 0.959 & 351.9 & 1.48 & 40.04 & 1675.7 & 11.23 \\
OmniControl~\citep{xie2024omnicontrol}
& \underline{0.877} & \underline{117.4}
& \underline{1.23} & \underline{37.12} & 1023.3 & 7.98
& \underline{0.877} & 130.4
& \underline{0.91} & \underline{30.73} & 622.6 & 5.18 \\
MotionLab~\citep{guo2025motionlab}
& 0.932 & 121.9 & 1.68 & 38.60
& \underline{760.9} & \underline{6.78}
& 0.915 & \underline{123.2} & 1.82 & 39.53
& \underline{388.1} & \underline{3.87} \\
\rowcolor{orange!12}
\textbf{MotionMaestro (Ours)}
& \textbf{0.841} & \textbf{82.4} & \textbf{0.75}
& \textbf{26.97} & \textbf{571.4} & \textbf{6.55}
& \textbf{0.760} & \textbf{69.9} & \textbf{0.59}
& \textbf{25.71} & \textbf{289.9} & \textbf{3.60} \\
\bottomrule
\end{tabular}%
}
\end{table*}

\paragraph{Effect of text guidance scale.}
Table~\ref{tab:cfg_main} varies text-only classifier-free guidance
for T2M, TP2M, and Long, while retaining non-text conditions in the
null branch. The default $w=1$ produces low FID but weaker text retrieval than
MotionLab, showing that favorable motion-distribution scores do
not by themselves establish stronger text alignment.
Moderate guidance can improve both quantities: for the default
model on MotionMillion, increasing $w$ from 1 to 2 reduces T2M FID
from 36.2 to 26.3 and raises R@1 from 0.401 to 0.554.
Further guidance generally increases retrieval scores but eventually
worsens FID, with the preferred balance depending on the task,
dataset, and model configuration. In particular, RoMo T2M FID for
MotionMaestro-5B is lowest at $w=1$ within the evaluated grid.

These are descriptive sweeps on test clips, not a validation-based
selection procedure; all our main comparisons remain at $w=1$.
The baselines retain their published guidance scales and are not
re-tuned. RoMo uses the MotionMillion-trained evaluator, so its
feature-space scores are interpreted within the same dataset and
protocol. Ground-truth retrieval scores provide reference values,
not upper bounds or a guarantee that generated motions exceeding
them are more faithful to real motion.

\begin{table*}[t]
\centering
\caption{
\textbf{Effect of text guidance scale $w$ on RoMo and MotionMillion.}
The null branch removes text while retaining all motion conditions.
$w=1$ ($^\dagger$) is used for all our main comparisons;
baselines retain their published scales.
\textbf{Bold} marks each model's lowest FID per task and dataset
across the grid, including ties. These test-set minima are not used
to select the main-table setting. Caption-free tasks are unaffected by $w$.
}
\label{tab:cfg_main}
\scriptsize
\setlength{\tabcolsep}{5pt}
\renewcommand{\arraystretch}{1.1}
\resizebox{\textwidth}{!}{%
\begin{tabular}{llcccccccccc}
\toprule
& & \multicolumn{5}{c}{\textbf{RoMo}}
  & \multicolumn{5}{c}{\textbf{MotionMillion}} \\
\cmidrule(lr){3-7}
\cmidrule(lr){8-12}
\textbf{Method} & $w$
& \makecell{\textbf{T2M}\\FID$\downarrow$}
& \makecell{\textbf{T2M}\\R@1$\uparrow$}
& \makecell{\textbf{TP2M}\\FID$\downarrow$}
& \makecell{\textbf{Long}\\FID$\downarrow$}
& \makecell{\textbf{Long}\\R@1$\uparrow$}
& \makecell{\textbf{T2M}\\FID$\downarrow$}
& \makecell{\textbf{T2M}\\R@1$\uparrow$}
& \makecell{\textbf{TP2M}\\FID$\downarrow$}
& \makecell{\textbf{Long}\\FID$\downarrow$}
& \makecell{\textbf{Long}\\R@1$\uparrow$} \\
\midrule
\rowcolor{black!5}
Ground truth & --
& -- & 0.363 & -- & -- & 0.370
& -- & 0.854 & -- & -- & 0.897 \\
\midrule
MDM & 2.5
& 450.7 & 0.205 & 456.1 & 510.9 & 0.220
& 297.6 & 0.323 & 334.4 & 336.2 & 0.281 \\
OmniControl & 2.5
& 199.4 & 0.338 & 312.8 & 398.6 & 0.278
& 153.9 & 0.484 & 193.8 & 177.9 & 0.493 \\
MotionLab & 5.75
& 101.5 & 0.424 & 46.5 & 123.5 & 0.380
& 61.4 & 0.706 & 49.4 & 108.6 & 0.662 \\
\midrule
\textbf{MotionMaestro} & 1.0$^\dagger$
& 14.9 & 0.293 & 28.7 & 26.2 & 0.258
& 36.2 & 0.401 & 52.8 & 46.4 & 0.436 \\
& 1.5
& \textbf{14.7} & 0.389 & \textbf{28.5} & 18.4 & 0.354
& 26.8 & 0.493 & 50.4 & 35.8 & 0.533 \\
& 2.0
& 18.0 & 0.446 & 29.4 & \textbf{17.3} & 0.414
& \textbf{26.3} & 0.554 & 48.5 & \textbf{33.5} & 0.585 \\
& 2.5
& 22.4 & 0.473 & 30.7 & 19.3 & 0.448
& 26.5 & 0.585 & \textbf{47.6} & \textbf{33.5} & 0.615 \\
& 3.0
& 26.7 & 0.493 & 32.2 & 22.5 & 0.466
& 28.0 & 0.609 & \textbf{47.6} & 34.1 & 0.641 \\
& 4.0
& 34.4 & 0.513 & 33.9 & 28.9 & 0.496
& 31.5 & 0.636 & 49.2 & 36.6 & 0.678 \\
& 5.0
& 40.9 & 0.524 & 38.1 & 34.2 & 0.501
& 35.1 & 0.669 & 50.7 & 39.6 & 0.694 \\
& 7.5
& 53.7 & 0.530 & 46.5 & 45.5 & 0.522
& 42.5 & 0.691 & 55.9 & 46.2 & 0.718 \\
\midrule
\textbf{MotionMaestro-5B} & 1.0$^\dagger$
& \textbf{17.2} & 0.309 & 28.6 & 19.0 & 0.279
& 34.7 & 0.362 & 42.8 & 35.8 & 0.407 \\
& 1.5
& 20.3 & 0.408 & 28.6 & \textbf{17.1} & 0.357
& \textbf{29.8} & 0.462 & 41.2 & \textbf{32.2} & 0.487 \\
& 2.0
& 24.1 & 0.450 & 28.5 & 20.0 & 0.411
& 31.0 & 0.525 & \textbf{41.0} & 33.1 & 0.544 \\
& 2.5
& 28.1 & 0.476 & \textbf{28.1} & 23.7 & 0.440
& 33.2 & 0.561 & 41.8 & 35.0 & 0.579 \\
& 3.0
& 31.4 & 0.488 & 28.8 & 27.6 & 0.463
& 35.7 & 0.586 & 42.0 & 37.4 & 0.605 \\
& 4.0
& 37.3 & 0.511 & 31.0 & 34.4 & 0.485
& 39.6 & 0.614 & 43.8 & 43.2 & 0.637 \\
& 5.0
& 41.6 & 0.517 & 32.5 & 40.1 & 0.505
& 43.0 & 0.639 & 45.2 & 47.2 & 0.662 \\
& 7.5
& 50.2 & 0.531 & 35.6 & 51.0 & 0.519
& 50.6 & 0.646 & 48.1 & 54.9 & 0.682 \\
\bottomrule
\end{tabular}%
}
\end{table*}

\section{Data and Preprocessing}
\label{app:mm_data}

We train a separate MotionMaestro model on each of RoMo and
MotionMillion. Unless otherwise specified, experiments use the default
MotionMaestro model and its three-stage training schedule.
Table~\ref{tab:supp_data} summarizes the data used after removing
numerically invalid clips. We train on the combined official training and validation splits and evaluate on the official test split. Reported results
use the final checkpoint rather than selecting a checkpoint by test
or validation performance.

\begin{table*}[htbp]
\centering
\caption{\textbf{Data and sequence lengths.}
Counts refer to the cleaned splits used in our experiments, not the
complete source releases. Standard evaluation lengths are the actual
exported lengths after deterministic bucketing and center cropping.}
\label{tab:supp_data}
\scriptsize
\setlength{\tabcolsep}{5pt}
\renewcommand{\arraystretch}{1.1}
\resizebox{0.7\textwidth}{!}{%
\begin{tabular}{lcc}
\toprule
\textbf{Setting} & \textbf{RoMo} & \textbf{MotionMillion} \\
\midrule
Training clips (train + validation) & 773,240 & 744,906 \\
Official test clips & 40,677 & 131,812 \\
Motion features per frame & 272 & 272 \\
Stage-3 maximum training length & 384 & 300 \\
Standard evaluation lengths (frames) & 24--128 & 56--256 \\
Long-sequence evaluation lengths (frames) & 129--300 & 129--300 \\
Evaluation caption & Level 1 & First paraphrase \\
\bottomrule
\end{tabular}%
}
\end{table*}

\paragraph{Motion preprocessing.}
Motions use 22 skeletal joints, metric coordinates, and a 30-fps
processing convention. Each crop is canonicalized to a common initial
root position and heading. Frame-zero displacement, heading increment,
and local velocity are reset to their boundary conventions.
Sequences are padded to a multiple of four; padding is excluded from
normalization statistics, attention keys, and loss reductions.
RoMo tokenizer training uses mirroring with probability $0.5$.
MotionMillion uses no additional mirroring, and neither dataset is
mirrored during generator training.

\paragraph{Text and normalization.}
RoMo training samples uniformly from caption levels 0--2;
MotionMillion samples from the available paraphrases. Captions are
used only for whole, unmirrored clips. Evaluation fixes the caption
selection in Table~\ref{tab:supp_data} across methods.
Feature normalization uses frozen, corpus-specific statistics with a
standard-deviation floor of $10^{-3}$. Nine structural constant channels
are excluded from reconstruction losses and restored at decoding.
Latent normalization is calibrated on 20,000 training-pool clips per
corpus, with a standard-deviation floor of $10^{-2}$.

\section{Architecture and Training Details}
\label{app:mm_implementation}

Unless explicitly stated otherwise, the tokenizer and training settings
below describe the default MotionMaestro model. Generator architecture
differences for MotionMaestro-5B are specified separately.

\paragraph{Representation and observation groups.}
Each frame contains $D=272$ features: 8 root-motion features,
$22\times3$ local joint positions, $22\times3$ local joint velocities,
and $22\times6$ joint rotations. The latent grid retains $J=23$ nodes: one root-motion node and 22 skeletal joints.

\paragraph{Tokenizer.}
The graph-temporal tokenizer has approximately 28.18M parameters.
Its embedding width is 128, and its four resolution levels use channel
widths $(128,256,512,512)$. The encoder uses two residual blocks per
level, temporal kernels of size three, and two stride-two temporal
downsampling operations. Spatial aggregation uses self connections,
skeletal edges, and a separate root--pelvis connection; there is no
joint-axis pooling. Encoder and decoder bottlenecks each use
8-head self-attention. The decoder uses three residual blocks per
level and has no encoder-to-decoder skip connections.
For padded length $\widetilde T$, the latent shape is
$K\times23\times16$, where $K=\widetilde T/4$.

\paragraph{Masking and condition encoding.}
Stage 1 combines anatomical tube, temporal stride, feature-group,
and short-window conditions with clean-motion reconstruction.
Anatomical tubes follow connected skeletal chains. Temporal anchors
use spacings from 2 to 8 frames, and feature-group conditions include
root-trajectory observations. Short inputs are encoded within their
cropped temporal support, rather than being required to reconstruct
an unspecified long continuation.
Stage 2 uses clean inputs and freezes the encoder.
At Stage 3, spatially or temporally distributed observations are
encoded on the masked target grid. Isolated poses, endpoint poses,
and prefixes are encoded on their own temporal support and placed on
the target latent grid. Locations without encoded support use the
learnable latent mask embedding. No-motion conditions bypass the
condition encoder.

\paragraph{Generator and observation map.}
MotionMaestro and MotionMaestro-5B contain approximately
1.29B and 5.10B generator parameters, respectively.
Both use 8 double-stream and 24 single-stream blocks,
24 attention heads, and an MLP expansion ratio of 3.
Their hidden widths are 1,536 and 3,072, yielding
per-head dimensions of 64 and 128, respectively.
Both use frozen FLAN-T5-XL features with width 2,048
and a maximum input length of 96 tokens.
MotionMaestro-5B additionally uses a 768-dimensional
pooled CLIP ViT-L/14~\citep{radford2021learning} text embedding with a maximum
input length of 77 tokens.
Learned node embeddings identify the 23 nodes.
Two-axis rotary embeddings encode text positions and
motion-frame coordinates, with head-dimension allocations
of $(16,48)$ for MotionMaestro and $(32,96)$ for
MotionMaestro-5B, both using base $10^4$.

\begin{table*}[htbp]
\centering
\caption{\textbf{Three-stage optimization settings.}
Settings for the default MotionMaestro model are shared across datasets
unless specified in the text.
Frame and token budgets are global upper budgets; the realized batch
size depends on sequence length.}
\label{tab:supp_training}
\scriptsize
\setlength{\tabcolsep}{5pt}
\renewcommand{\arraystretch}{1.1}
\resizebox{0.75\textwidth}{!}{%
\begin{tabular}{lccc}
\toprule
\textbf{Setting} & \textbf{Stage 1} & \textbf{Stage 2} & \textbf{Stage 3} \\
\midrule
Trainable components & Encoder + decoder & Decoder & Generator \\
Updates & 100,000 & 20,000 & 200,000 \\
Peak learning rate & $10^{-4}$ & $2\times10^{-5}$ & $10^{-4}$ \\
Learning-rate warmup & 2,000 & 500 & 2,000 \\
Final / peak learning rate & 0.1 & 0.1 & 0.1 \\
Global padded-frame budget & 49,152 & 49,152 & 8,192 \\
Global generator-token budget & -- & -- & 54,000 \\
Gradient accumulation & 1 & 1 & 1 \\
Optimizer & AdamW & AdamW & AdamW \\
AdamW $(\beta_1,\beta_2)$ & $(0.9,0.95)$ & $(0.9,0.95)$ & $(0.9,0.95)$ \\
Weight decay & 0.01 & 0.01 & 0.01 \\
Gradient-norm clipping & 1.0 & 1.0 & 1.0 \\
Precision & bfloat16 & bfloat16 & bfloat16 \\
GPUs & $2\times$B200 & $2\times$B200 & $2\times$B200 \\
\bottomrule
\end{tabular}%
}
\end{table*}

\paragraph{Length sampling and losses.}
Tokenizer training samples ordinary windows of 24--128 frames at
8-frame intervals, longer windows of 192 and 256 frames, and a final
long-window length of 384 on RoMo or 300 on MotionMillion. It also
includes short windows of 1--16 frames.
Generator training retains whole clips within the dataset-specific
maximum and randomly crops longer clips.
Training settings are listed in Table~\ref{tab:supp_training}.
After warm-up, cosine schedules reduce the learning rates to
one-tenth of their peaks. Variable-length clips are batched using
length buckets and the global budgets in the table.
Stage~3 freezes the entire tokenizer. Generator training takes
approximately 48 hours per dataset.
We use the objectives defined in the main paper, with
$\lambda_p=\lambda_y=0.1$ and maximum $\lambda_o=0.1$.
The observation-loss weight increases linearly from zero during the
first 5,000 generator updates.
Feature-space losses average valid entries within each sample and
active feature families before averaging samples. Examples without motion
observations contribute zero observation loss.

\paragraph{Task and text sampling.}
The eight training families have probabilities 0.25 for T2M, 0.05 for
unconditional generation, 0.15 for partial completion, 0.20 for
keyframe conditioning, 0.15 for trajectory conditioning, 0.05 for
P2M, 0.05 for TP2M, and 0.10 for continuation.
FLF and FI are keyframe presets, while Long evaluates text-conditioned
generation at longer lengths, yielding the ten evaluation tasks.
Text is supplied to T2M and TP2M, withheld for unconditional generation
and P2M, and sampled with probability $0.5$ for the remaining training
families when an aligned caption is available.
Text dropout is $0.1$ and retains all motion observations.
When no aligned caption exists, T2M and TP2M are excluded and the
remaining task probabilities are renormalized.

\paragraph{Sampling.}
We follow the main-paper flow convention,
$\mathbf z_t=(1-t)\mathbf z_0+t\boldsymbol\epsilon$, and integrate
from $t=1$ to $t=0$. With $N=50$ Euler steps and $t_i=1-i/N$,
\begin{equation}
\mathbf z_{t_{i+1}}=\mathbf z_{t_i}
-\frac{1}{N}\,\mathbf v_\theta
(\mathbf z_{t_i},t_i;\mathbf z^c,\mathbf m,\mathbf c).
\end{equation}
Our main-table results use final EMA weights with decay $0.999$,
sampling seed 0, and text guidance scale $w=1$.
For guided sampling,
$\mathbf v^{(w)}=\mathbf v_{\varnothing}
+w(\mathbf v_{\mathbf c}-\mathbf v_{\varnothing})$;
the null branch removes only text and keeps the motion condition.
The 50-step sampler requires 50 function evaluations at $w=1$ and
100 when both guidance branches are evaluated.
Outputs are decoded without copying observations back, latent
clamping, smoothing, inverse kinematics, or foot-contact correction.

\section{Evaluation Protocol and Baseline Adaptations}
\label{app:mm_protocol}

Table~\ref{tab:supp_tasks} defines the ten main evaluation tasks.
The stride and body-part experiments change only the specified
observation presets within their respective evaluation protocols.

\begin{table*}[htbp]
\centering
\caption{\textbf{The ten main-paper evaluation tasks.}
A pose includes joint positions and rotations together with root
placement and heading. Text is provided only for T2M, TP2M, and Long.
MPJPE is evaluated on unobserved joint positions; FID uses the whole
output sequence.}
\label{tab:supp_tasks}
\scriptsize
\setlength{\tabcolsep}{5pt}
\renewcommand{\arraystretch}{1.1}
\resizebox{0.8\textwidth}{!}{%
\begin{tabular}{llcc}
\toprule
\textbf{Task} & \textbf{Motion observation} & \textbf{Text} & \textbf{Main metric} \\
\midrule
Uncond. & None & No & FID \\
T2M & None & Yes & FID \\
P2M & First pose & No & FID \\
TP2M & First pose & Yes & FID \\
FLF & First and last poses & No & FID \\
FI & Periodic poses at spacing $s=4$ & No & Hidden MPJPE \\
Extrap. & A motion prefix & No & FID \\
Partial & Positions of 13 upper-body joints at all frames & No & Hidden MPJPE \\
Trajectory & Root $xz$ at all frames; no heading or joint poses & No & FID \\
Long & None; 129--300 output frames & Yes & FID \\
\bottomrule
\end{tabular}%
}
\end{table*}

\paragraph{Metrics.}
FID uses 512-dimensional motion features from the frozen MotionMillion
evaluator. Inputs are padded to 300 frames, normalized using the
evaluator's own statistics, and embedded using the posterior mean.
The same evaluator is used for both datasets; RoMo is out of domain,
so its feature-space scores should be interpreted within this protocol.
FID is computed against the corresponding reference set. Scores from
different tasks or sample counts are not directly comparable.

MPJPE measures Euclidean error in heading-local joint positions,
in millimeters, excluding padding and observed position cells.
Root-path error measures Euclidean $xz$ error at supplied path frames.
For text evaluation, R-Precision uses identical motion--caption pairs
across methods; matching distance is the mean distance between paired
text and motion embeddings.

\paragraph{Baseline adaptations.}
MDM~\citep{tevet2022mdm}, OmniControl~\citep{xie2024omnicontrol}, and
MotionLab~\citep{guo2025motionlab} are trained on the target datasets
and scored with a common evaluator.
MDM uses an 8-layer, width-512 transformer, batch size 64, learning
rate $10^{-4}$, and 1,000-step diffusion. It uses inpainting for
supported observations and cannot directly receive an absolute
root-$xz$ trajectory.
OmniControl is fine-tuned for 400,000 updates at learning rate
$10^{-5}$ from our corpus-matched MDM, using batch size 64 and
1,000-step diffusion with spatial guidance. Its adapter accepts only
joints in the published six-joint control set; upper-body hints are
restricted to the head and two hands.
MotionLab uses batch size 256, learning rate $10^{-4}$, and 242
training epochs, followed by 50 Euler sampling steps.
Text-guided baseline sampling uses scales 2.5 for MDM and OmniControl
and 5.75 for MotionLab; caption-free presets use their null-text
settings.
These are corpus-matched evaluations with a common scorer, not
comparisons under matched training or inference compute budgets.

\clearpage

\bibliography{iclr2027_conference}
\bibliographystyle{iclr2027_conference}

\end{document}